\documentclass[lettersize,journal]{IEEEtran}
\usepackage{amsmath,amsfonts}
\usepackage{algorithm}
\usepackage{algorithmicx}
\usepackage{array}
\usepackage[caption=false,font=normalsize,labelfont=sf,textfont=sf]{subfig}
\usepackage{textcomp}
\usepackage{stfloats}
\usepackage{url}
\usepackage{verbatim}
\usepackage{graphicx}
\usepackage{bbm}
\usepackage{yhmath}
\usepackage{multirow}

\usepackage[colorlinks,
            linkcolor=blue,
            anchorcolor=blue,
            citecolor=blue,
            urlcolor=blue]{hyperref}
\def\BibTeX{{\rm B\kern-.05em{\sc i\kern-.025em b}\kern-.08em
    T\kern-.1667em\lower.7ex\hbox{E}\kern-.125emX}}
\usepackage{balance}
\begin{document}
\title{\textcolor{blue}{Scene Modeling of Autonomous Vehicles Avoiding Stationary and Moving Vehicles on Narrow Roads}}

\author{Qianyi Zhang, Jinzheng Guang, Zhenzhong Cao, and Jingtai Liu,~\IEEEmembership{Senior Member, IEEE}
\thanks{The authors are with the Institute of Robotics and Automatic Information System, Nankai University, Tianjin 300350, China; 
Tianjin Key Laboratory of Intelligent Robotics, Tianjin 300350, China; 
and also with TBI center, Nankai University, Tianjin 300350, China. 
Emails of the first author: zhangqianyi@mail.nankai.edu.cn; corresponding author: liujt@nankai.edu.cn. 
This work is supported by National Natural Science Foundation of China under Grant 62173189.}}

\markboth{}%
{How to Use the IEEEtran \LaTeX \ Templates}

\maketitle


\begin{abstract}
Navigating narrow roads with oncoming vehicles is a significant challenge that has garnered considerable public interest. These scenarios often involve sections that cannot accommodate two \textcolor{blue}{moving vehicles} simultaneously due to the presence of \textcolor{blue}{stationary vehicles} or limited road width. Autonomous vehicles must therefore profoundly comprehend their surroundings to identify passable areas and execute sophisticated maneuvers. 
To address this issue, this paper presents a comprehensive model for such an intricate scenario. The primary contribution is the principle of road width occupancy minimization, which models the narrow road problem and identifies candidate meeting gaps. Additionally, the concept of homology classes is introduced to help initialize and optimize candidate trajectories, while evaluation strategies are developed to select the optimal gap and most efficient trajectory. 
Qualitative and quantitative simulations demonstrate that the proposed approach, SM-NR, achieves high scene pass rates, efficient movement, and robust decisions. 
Experiments conducted in tiny gap scenarios and conflict scenarios reveal that the autonomous vehicle can robustly select meeting gaps and trajectories, compromising flexibly for safety while advancing bravely for efficiency. 
Visit \href{https://sm-nr.github.io}{https://sm-nr.github.io} for the video and code.

\textcolor{red}{This work has been submitted to the IEEE for possible publication. Copyright may be transferred without notice, after which this version may no longer be accessible.}

\end{abstract}


\section{Introduction}

\IEEEPARstart{W}{ith} the advancement of navigation systems, the autonomous vehicles have developed from controlled laboratory scenarios to everyday roads~\cite{2023_tits_2, 2024_TSMC_1, 2024_TSMC_4}. 
While autonomous navigation is generally safe and efficient, collision risks or traffic congestion still occasionally occur, especially in complex scenarios such as narrow roads~\cite{IROS_ZHANG_2022}, intersections~\cite{intro_2022_tits_1}, and expressways~\cite{2023_tits_3}. 
To address these issues, current navigation systems have evolved into multi-model frameworks~\cite{2022_nr_1}, 
employing a robust general navigation method most of the time while switching to specialized branches in response to specific scenarios. 
Consequently, identifying challenging scenarios and developing corresponding effective solutions for autonomous vehicles constitutes a critical and practical research area~\cite{2023_tits_4}.

\begin{figure}[htb]
	\centering
    \includegraphics[width=3.5in]{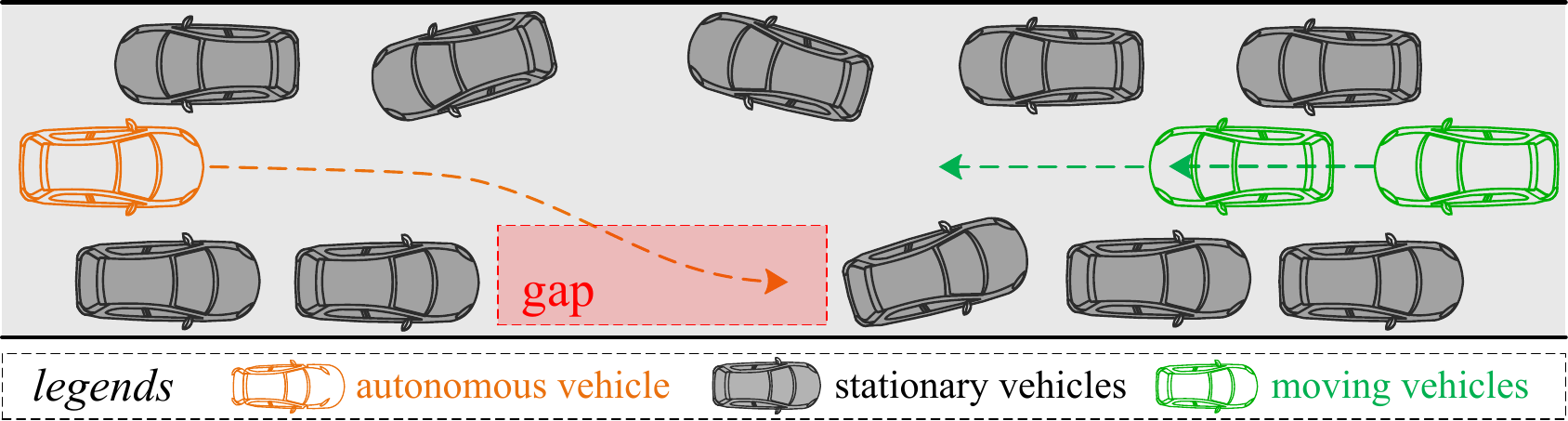}
	\caption{\textcolor{blue}{Illustration of vehicles passing through the narrow road.} 
    The autonomous vehicle is expected to proactively pull over and wait at the red gap to avoid \textcolor{blue}{moving vehicles approaching from the oncoming direction}. 
    }
	\label{fig_motivation}
\end{figure}

In the context of challenging driving scenarios, navigating narrow roads is a frequent challenge in residential communities and rural areas~\cite{2021_nr_1}. 
A road is considered a narrow road when part of the road cannot allow two moving vehicles to meet, either due to the road's limited width or the presence of \textcolor{blue}{stationary vehicles} along the roads. 
When two vehicles approach from opposite directions on such a road, one must proactively yield to prevent blocking the road. 
This bottleneck situation~\cite{2023_nr_1,2023_tits_5} is exemplified in Fig.\ref{fig_motivation}, where an autonomous vehicle encounters road blockage caused by \textcolor{blue}{stationary vehicles} and other \textcolor{blue}{moving vehicles}. 
In such situations, traditional navigations often struggle to identify a viable trajectory, resulting in the autonomous vehicle oscillating indecisively in place and awkwardly anticipating \textcolor{blue}{moving vehicles} to either advance into the red gap or reverse to the rightmost area to resolve the conflict. 
However, when the red gap is not long enough to accommodate all the two \textcolor{blue}{moving vehicles}, the optimal strategy is that the autonomous vehicle proactively pulls over at the gap and waits until the \textcolor{blue}{moving vehicles} pass it~\cite{2020_nr_2}. 
Such proactive maneuvers are critical for improving the efficiency and safety of autonomous navigation on narrow roads. 

To solve this problem, the primary challenge is to identify gaps that can simultaneously accommodate two vehicles approaching from opposite directions. Unlike normal roads, passible gaps on narrow roads are often irregularly located in very small patches between several vehicles. Furthermore, even though some areas are large enough, they will still be further filtered by vehicle kinematic constraints, as these areas must be accessible by the vehicles. 
Even though some previous work~\cite{2021_nr_1, 2020_nr_2, 2015_nr_1, 2022_nr_4, 2024_TSMC_3, 2024_TSMC_5} used simple narrow roads as one of the test scenarios, none of them refined the characteristics of this scenario. 
To address this gap, our work builds a fine model for the narrow road scenario for the first time (see Sec.\ref{3B}-\ref{3D}). 
Furthermore, a principle of road width occupancy minimization is designed to segment the environment and identify the meeting gaps according to the distribution of \textcolor{blue}{stationary vehicles} (see Sec.\ref{3E}). 
By assuming the vehicles approaching from two ends advance tightly alongside the \textcolor{blue}{stationary vehicles} and road edges, two expanded boundaries are formed. 
According to whether the two expanded boundaries intersect, the whole road is divided into non-meeting areas and meeting gaps which provide enough room for vehicles approaching from opposite ends to meet here. 
Given these candidate meeting gaps, it is essential to determine an optimal gap based on the relative positions and speeds of the autonomous vehicle and other \textcolor{blue}{moving vehicles}. To achieve this, a minimal model is proposed to categorize potential situations and evaluate the meeting gaps (see Sec.\ref{3F}).

Given an identified optimal gap, the concept of the homology class~\cite{H_signature} is introduced to assist the autonomous vehicle in initializing trajectories with distinct meanings, such as cutting into the gap or backing into it (see Sec.\ref{4A}). Each trajectory undergoes individual optimization to meet the vehicle’s kinematic constraints, collision avoidance requirements, and the consistency of its meaning (see Sec.\ref{4B}). The optimized trajectories are then evaluated by a hierarchical evaluation strategy to select the optimal one that best balances the efficiency and safety of the autonomous vehicle(see Sec.\ref{4C}).

\begin{figure*}[htb]
	\centering
    \includegraphics[width=7.1in]{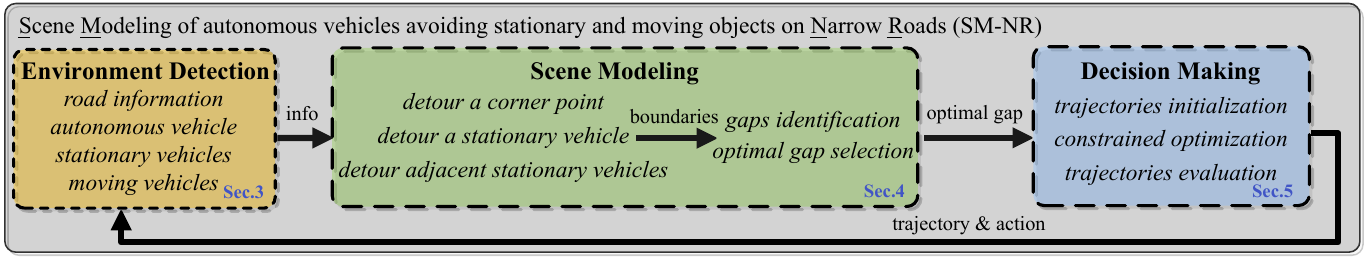}
	\caption{\textcolor{blue}{Framework of our proposed method, SM-NR.} Taking road information and vehicle distribution as input, boundaries are outlined, candidate meeting gaps are identified, and the optimal gap is selected. Candidate trajectories are then initialized and optimized, with the optimal one selected and executed.
    }
	\label{fig_framework}
\end{figure*}


In summary, this work builds a model for the narrow road scenario for the first time, and presents a navigation framework for autonomous vehicles to efficiently pass through narrow roads. 
Three contributions are as follows: 
\begin{itemize}
  \item A principle of road width occupancy minimization is designed to model the narrow road and identify gaps. 
  \item A concept of homology class is introduced to initialize and optimize candidate trajectories for meeting gaps. 
  \item Two evaluation strategies are developed to select the optimal gap as well as the optimal trajectory. 
\end{itemize}

The remainder of the paper is organized as follows:
Sec.\ref{sec:2} reviews related work. 
Sec.\ref{assumpt} presents the \textcolor{blue}{assumptions and problem formulations}.
Sec.\ref{sec:3} details the scene model for narrow roads.
Sec.\ref{sec:4} outlines the process of trajectory initialization, optimization, and evaluation.
Sec.\ref{sec:5} validates our approach through qualitative and quantitative simulations.
Sec.\ref{sec:6} assesses the performance of our approach in real-world scenarios.
Finally, Sec.\ref{sec:8} concludes the paper and discusses its limitations.

\section{Related Works} \label{sec:2}
\textcolor{blue}{With the growing commercial interest in mobile robots and autonomous driving, research efforts have increasingly focused on scene modeling and navigation in complex environments~\cite{2022_nr_1, scene1, scene2, scene3, scene4}. Typical frameworks leverage LiDAR or camera data to estimate road boundaries, traffic signs, vehicle positions, and vehicle future trajectories. Based on this information, the autonomous vehicle’s decisions are made, trajectories are planned, and actions are executed.}

\textcolor{blue}{One prominent research direction within this field explores human-robot interaction in the spatial-temporal dimension.} Notable studies~\cite{2011_nr_1, 2014_nr_1} project static obstacles and dynamic pedestrians into an X-Y-T space, under the assumption that pedestrians maintain their current velocities. Trajectories are subsequently generated and evaluated according to personal space, social norms, and inferred human intentions. In parallel, Khambhaita and colleagues~\cite{2020_nr_2, 2017_nr_1, 2020_nr_1} have focused on human-robot collaboration, anticipating mutual reactions between humans and robots. In this context, robot decisions are optimized by incorporating predictions of human trajectories alongside robotic constraints. An especially notable contribution is the planning framework proposed by Che~\cite{2020_nr_3}, which integrates both implicit (e.g., motion constraints) and explicit (e.g., gaze, visual, and auditory cues) communication, enabling multi-modal decision-making in human-robot interactions.

\textcolor{blue}{Another line of research has introduced preset semantic strategies for robotic actions, such as reversing, merging, and turning~\cite{2015_nr_2, 2015_nr_3}.} These strategies are assessed based on the spatial relationship between the robot and surrounding obstacles, with the optimal strategy selected and converted into corresponding maneuvers or trajectories. Furthermore, several multi-stage frameworks~\cite{2022_nr_4, 2022_nr_3, 2022_nr_2} have been proposed to balance potential strategies, optimize trajectories, and evaluate them through cost functions. These cost functions consider factors such as predicted human movement, reactive responses, active motion-inducing interactions, spatial proximity, relative speeds, group sizes, and interactions among group members.

\textcolor{blue}{A third research branch investigates multi-agent interactions involving multiple moving vehicles~\cite{2021_nr_2}.} Here, human personalities and cooperative intentions are accounted for in the interaction modeling. The chicken game~\cite{2013_game_1}, a classic zero-sum competitive game, illustrates situations where players compete for maximum gain and are disinclined to yield. Conversely, the cooperative game~\cite{2020_nr_2} emphasizes players’ preference to cooperate, thereby avoiding adverse outcomes like collisions. To improve robustness, Liu~\cite{game_1} proposed a framework that incorporates two potential games, providing theoretical guarantees for the existence of pure-strategy Nash equilibria, thereby accelerating convergence to equilibrium and enhancing decision robustness. For safety, Yang~\cite{game_2} introduced a prediction-planning collaboration method to help autonomous vehicles avoid collisions with moving pedestrians by modeling pedestrian-vehicle interactions in shared spaces. To increase computational efficiency, Zanardi~\cite{game_3} factorized the dynamic game to leverage vehicle independence, yielding a streamlined graph structure and improved efficiency. Finally, considering drivers’ preferences, Chandra~\cite{game_4} developed the CMetric, a tool for learning drivers’ risk preferences from datasets, enabling risk-aware modeling of interactions between social and autonomous vehicles.

\textcolor{blue}{Although the works mentioned above have introduced many interesting ideas and perform well in various scenarios, they are only effective in the simplest narrow-road situations. In more complex cases, such as the one illustrated in Fig.\ref{fig_motivation}, where moving objects approach from the opposite direction, they often come to an awkward stop, ultimately causing traffic congestion. This limited performance stems from overlooking the unique demands of the narrow road scenario, which requires autonomous vehicles to proactively execute additional maneuvers, such as moving close to the road edge or parking adjacent to other vehicles~\cite{2022_nr_5}. Safely performing these actions requires a comprehensive analysis of the static environment, including road features, the distribution of stationary vehicles, and the states of other moving vehicles.}

Peter~\cite{2010_nr_1} summarized this issue as the ``freezing robot problem'', wherein robots perceive all forward paths as unsafe when the environmental complexity exceeds a certain threshold. This leads to robots freezing in place or performing unnecessary maneuvers, such as back-and-forth movements or oscillations, to avoid potential collisions. 
\textcolor{blue}{Three studies, most relevant to this work, address this issue from different perspectives. Wang~\cite{2021_nr_1} simplifies LiDAR data into a spatial grid map, using convolutional neural networks and reinforcement learning to support autonomous driving decisions, though it does not consider road boundaries or model interactions between vehicles. Ippei~\cite{2015_nr_1} estimates object speeds in the environment and predicts future trajectories to identify collision-free paths in the temporal-spatial dimension, accounting for basic environmental information but lacking cooperative intent between vehicles. P2EG~\cite{IROS_ZHANG_2022} introduces a tentative game to model vehicle interactions and reach a mutually accepted solution, yet it overlooks environmental features.}

\textcolor{blue}{Building on these studies, this work presents a more comprehensive and detailed model for narrow-road meeting scenarios. It offers a refined consideration of road structure, stationary and moving vehicles, and the kinematic model of the autonomous vehicle, introducing the concept of meeting gaps and enabling their precise identification. Additionally, it incorporates the interaction and cooperation between autonomous and moving vehicles to identify the optimal gap and trajectory.}

\begin{figure*}[htb]
\centering
\includegraphics[width=7.1in]{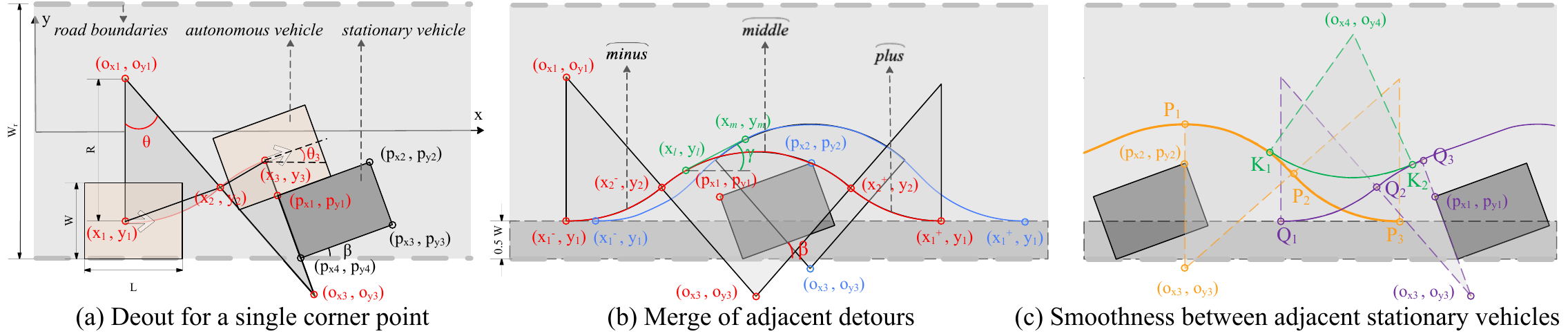}
\caption{\textcolor{blue}{Illustration of scene modeling for detouring a stationary vehicle.} 
The autonomous vehicle is assumed to move closely alongside the road boundary and \textcolor{blue}{stationary vehicles}.
(a) A \textcolor{blue}{stationary vehicle} is simplified as four corner points $(p_{xi}, p_{yi})|_{i=\{1,2,3,4\}}$. To detour around a point $(p_{x1}, p_{y1})$, the autonomous vehicle starts to turn left at $(x_1,y_1)$ to leave the road boundary and then turns right at $(x_2,y_2)$ until it is in parallel with the \textcolor{blue}{stationary vehicle} at $(x_3,y_3)$. 
(b) A whole detour around $(p_{x1}, p_{y1})$ is illustrated in red, consisting of three parts: $minus$, $middle$, and $plus$. Another detour around $(p_{x2},p_{y2})$ in blue intersects with the red one, and they are merged smoothly with a green connection with an orientation $\gamma$, to form a complete detour around the entire \textcolor{blue}{stationary vehicle}. 
(c) When two \textcolor{blue}{stationary vehicles} are close, their detours are smoothly connected with a green curve, which is part of a circle centered at $(o_{x4},o_{y4})$. 
}
\label{fig_environment_segmentation}
\end{figure*}

\section{\textcolor{blue}{Assumption and Problem Formulation}} \label{assumpt}
\textcolor{blue}{In the narrow road scenario, the autonomous vehicle, located at $(x, y)$ with orientation $\theta$, length $L$, and width $W$, observes its surroundings with LiDAR and cameras and plans a trajectory $\tau$ to advance. Note that the center of the rear wheels of the autonomous vehicle is used to represent its location.}

\textcolor{blue}{During this process, the detection and tracking module further extracts the observation as boundaries $\mathbf{B}_{upper}$ and $\mathbf{B}_{lower}$, stationary vehicles $\mathbb{S}$, and moving vehicles $\mathbb{M}$ under the SL coordinate system~\cite{frenet1, frenet2}. The SL coordinate system is oriented with the x-axis aligned along the road centerline in the vehicle’s forward direction, and the y-axis perpendicular to x, forming a right-handed system.}

\textcolor{blue}{The road boundaries refer to the edges of the road, where their coordinates along x are denoted as $y_{up} = \mathbf{B}_{up}(x)$ and $y_{down} = \mathbf{B}_{down}(x)$. The subscripts ``up'' and ``down'' indicate the side with greater and smaller y-values, respectively. These boundaries are typically represented as polynomial curves. 
In this paper, we assume the road boundaries are straight, such that $\mathbf{B}_{up}(x) = W_r/2$ and $\mathbf{B}_{down}(x) = -W_r/2$ for any x, where $W_r$ is the detected road width. }

\textcolor{blue}{Each Stationary vehicle $\textbf{s}_i \in \mathbb{S}$, parked along the boundary with orientation $\beta_i$, is simplified as a rectangle with four corner points $\textbf{s}_i = \{ \textbf{p}_{1}, \textbf{p}_{2}, \textbf{p}_{3}, \textbf{p}_{4} \}$ with each element $\textbf{p}_{j} = (p_{xj}, p_{yj})$ to differentiate symbols with the autonomous vehicle.}

\textcolor{blue}{Each moving vehicle $\textbf{m}_i \in \mathbb{M}$, approaching from the oncoming direction of the autonomous vehicle, is represented similarly to the autonomous vehicle but is distinguished by an apostrophe in its notation, e.g., $(x', y', \theta')$.}

\textcolor{blue}{Based on this information, several candidate meeting gaps $\mathbb{G}$ are identified that are wide enough for the autonomous and moving vehicles to pass each other. The optimal gap $g^* \in \mathbb{G}$ is then selected, and several candidate trajectories $\mathbb{T}$ are generated to guide the autonomous vehicle into this gap. After evaluating all the trajectories, the optimal one $\tau^* \in \mathbb{T}$ is chosen and executed by the autonomous vehicle. After yielding to the moving vehicle by stopping in the gap, the autonomous vehicle follows a recovery trajectory toward a local goal positioned a few meters ahead along the road centerline, enabling it to resume normal advancement and prepare for the next meeting.}

\section{Scene Modeling for Narrow Road} \label{sec:3}
This section outlines the process of identifying an optimal gap. The first three subsections assume the autonomous vehicle moves along the road and \textcolor{blue}{stationary vehicles}, forming an expanded boundary based on the principle of road width occupancy minimization. The subsequent subsections categorize the road into meeting gaps and non-meeting areas and select the optimal gap for the autonomous vehicle.

\subsection{Detour a Single Corner Point of \textcolor{blue}{Stationary Vehicle}} \label{3B}
This subsection demonstrates how the autonomous vehicle detours a corner point \textcolor{blue}{$(p_{xi}, p_{yi})$} of the \textcolor{blue}{stationary vehicle}, resulting in a red trajectory, as illustrated in Fig.\ref{fig_environment_segmentation}(b). This trajectory comprises three alternative turns $minus$, $middle$, and $plus$. The focus is to solve for the four endpoints of these turns: \textcolor{blue}{$(x_1^-, y_1),  (x_2^-, y_2),  (x_2^+, y_2), and (x_1^+, y_1)$}.

The solving details are illustrated in Fig.\ref{fig_environment_segmentation}(a). The turn $minus$ is part of a circle centered at \textcolor{blue}{$(o_{x1}, o_{y1})$} with a radius equal to the autonomous vehicle’s minimum turning radius $R$, beginning at \textcolor{blue}{$(x_1, y_1)$} and ending at a switch point \textcolor{blue}{$(x_2, y_2)$} by a degree $\theta$. The turn $middle$ is also part of a circle, centered at \textcolor{blue}{$(o_{x3}, o_{y3})$} with radius $R$, starting from \textcolor{blue}{$(x_2, y_2)$} and passing through a specific point \textcolor{blue}{$(x_3, y_3, \theta_3)$}.
The known information is that the starting position is parallel to and adjacent to the road boundary, meaning that \textcolor{blue}{$y_1$} is known and offset from the road boundary by half of the vehicle width $W/2$. 
    
The definition of the specific point \textcolor{blue}{$(x_3, y_3, \theta_3)$} is based on the corner point \textcolor{blue}{$(p_{xi}, p_{yi})$}. It has a position \textcolor{blue}{$(x_3, y_3)$} that is the closest to the corner point and an orientation \textcolor{blue}{$\theta_3$} that depends on the corner point's relative position to other corner points. For the corner point closest to the road centerline, the orientation is parallel to the road boundary with \textcolor{blue}{$\theta_3=0$}; otherwise, the orientation aligns with the body of the \textcolor{blue}{stationary vehicle}. Taking the \textcolor{blue}{stationary vehicle} in Fig.\ref{fig_environment_segmentation}(a) as an example, \textcolor{blue}{$\theta_{3}(p_{x1}, p_{y1}) = \arctan(p_{y2} - p_{y1}, p_{x2} - p_{x1})$, $\theta_{3}(p_{x2}, p_{y2}) = 0$, and $\theta_{3}(p_{x3}, p_{y3}) = \arctan(p_{y3} - p_{y2}, p_{x3} - p_{x2})$}. 

Given this information and definitions, the variable \textcolor{blue}{$(x_2, y_2)$} and $\theta$ can be solved in terms of the known parameters: the corner point \textcolor{blue}{$(p_{xi}, p_{yi})$}, vehicle shape $W$ and $L$, turning radius $R$, and road width $W_r$.
For \textcolor{blue}{$y_2$}: 
\begin{equation}
  \begin{split}
    & y_1 = -W_r / 2 + W/2\\
    & p_{yi}-o_{y3}=(R-W/2)cos\theta_3 \\
    & o_{y1}=y_1+R \\
    & y_2=(o_{y1}+o_{y2})/2
  \end{split}
\end{equation}
For $\theta$ and \textcolor{blue}{$x_2$}: 
\begin{equation}
  \begin{split}
    & Rcos\theta=o_{y1}-y_1 \\
    & o_{x3}-p_{xi} = (R-W/2)sin\theta_3 \\
    & (x_2-o_{x3})^2+(y_2-o_{y3})^2=R^2
  \end{split}
\end{equation}
The solutions are: 
\begin{equation}
  \begin{split}
    & x_2^- = p_{xi}+(R-W/2)sin\theta_3 - \sqrt{R^2-(A-B)^2}/2 \\
    & x_2^+ = p_{xi}+(R-W/2)sin\theta_3 + \sqrt{R^2-(A-B)^2}/2 \\
    & y_2 = (A+B)/2 \\
    & \theta = |arccos( (A-B) / (2R) )|
  \end{split}
\end{equation}
where $A = R-W_r / 2 + W/2$ and $B = p_{yi}-(R-W/2)cos\theta_3$. 
Note that, due to the symmetry, both \textcolor{blue}{$x_2^-$} and \textcolor{blue}{$x_2^+$} can be obtained directly. These variables are illustrated in Fig.\ref{fig_environment_segmentation}(b). 

Consider the turnings $minus$ and $plus$ on both sides: 
\begin{equation}
    x_2-x_1=Rsin\theta
\end{equation}
The start and final states \textcolor{blue}{$x_{1}^-$}, \textcolor{blue}{$x_{1}^+$} are solved as: 
\begin{equation}
  \begin{split}
    & x_1^- = p_{xi} + Rsin\theta_3 - Rsin|\theta| -\sqrt{R^2-(A-B)^2}/2  \\
    & x_1^+ = p_{xi} + Rsin\theta_3 + Rsin|\theta| +\sqrt{R^2-(A-B)^2}/2  
  \end{split}
\end{equation}


\subsection{Merge of Adjacent Detours of a \textcolor{blue}{Stationary Vehicle}}
This subsection demonstrates how to merge two detours around adjacent corner points belonging to the same \textcolor{blue}{stationary vehicle} to form a smooth detour. 
A \textcolor{blue}{stationary vehicle} contains up to three valid corner points. 
\textcolor{blue}{For clarity, we temporarily add distinct footnotes to variables in this subsection, with $l$, $m$, and $r$ representing the variables associated with the leftmost, middle, and rightmost corner points, respectively. 
Fig.\ref{fig_environment_segmentation}(b) illustrates the variables with footnote $l$ in red, variables with footnote $m$ in blue, and the connection between them in green. }

When two detours around adjacent corner points intersect, the gap between them is smoothly filled to ensure the merged detour satisfies the kinematic constraints of the autonomous vehicle. Recall that any detour consists of three parts $minus$, $middle$, and $plus$, and a \textcolor{blue}{stationary vehicle} has at most three detours $l$, $m$, and $r$. 
As the circle center's y-coordinate for both $minus$ and $plus$ in any detour is the same value $o_{y1}$, indicating that the autonomous vehicle starts from and returns to a state next to the road boundary, 
whether two adjacent detours intersect can be identified by comparing their ends: 
\begin{equation}
  \begin{split}
    if \; x^-_{1,l} < x^-_{1,m} \; & \Rightarrow \; intersect \; (l \& m) \\
    if \; x^+_{1,m} < x^+_{1,r} \; & \Rightarrow \; intersect \; (m \& r) 
  \end{split}
\end{equation}

For two intersected detours, the intersection must occur at their middle turns $middle$. Assuming the two turns reach the same gradient $\gamma$ at respectively \textcolor{blue}{$(x_{l},y_{l})$} and \textcolor{blue}{$(x_{m},y_{m})$} around the intersection point, they are represented as: 
\begin{equation}
  \begin{split}
    & x_{l} = o_{x3,l} - Rsin\gamma, \hspace{6pt} y_{l} = o_{y3,l} + Rcos\gamma \\
    & x_{m} = o_{x3,m} - Rsin\gamma, \hspace{6pt} y_{m} = o_{y3,m} + Rcos\gamma \\
    & tan\gamma=(y_{l}-y_{m})/(x_{l}-x_{m})
  \end{split}
\end{equation}
The solutions are: 
\begin{equation}
  \begin{split}
  & \gamma = \frac{ (o_{y3,l}-(R-\frac{W}{2})cos\theta_{3,l}) - (o_{y3,m}-(R-\frac{W}{2})cos\theta_{3,m}) }{ (o_{x3,l}+(R-\frac{W}{2})sin\theta_{3,l}) - (o_{x3,m}+(R-\frac{W}{2})sin\theta_{3,m})} \\  
    & x_{l} = (o_{x3,l}+(R-W/2)sin\theta_{3,l}) -Rsin\gamma \\
    & y_{l} = (o_{y3,l}-(R-W/2)cos\theta_{3,l}) + Rcos\gamma \\
    & x_{m} = (o_{x3,m}+(R-W/2)sin\theta_{3,m}) -Rsin\gamma \\
    & y_{m} = (o_{y3,m}-(R-W/2)cos\theta_{3,m})  + Rcos\gamma
  \end{split}
\end{equation}
In this case, the connection of all detours forms a complete detour that enables the autonomous vehicle to move around the whole \textcolor{blue}{stationary vehicle} smoothly.

\subsection{Smoothness between Adjacent \textcolor{blue}{Stationary Vehicles}} \label{3D}
When two \textcolor{blue}{stationary vehicles} are far apart, there is no intersection between their detours. The autonomous vehicle can overtake one \textcolor{blue}{stationary vehicle}, return next to the road, and then overtake the other. However, when the space between two \textcolor{blue}{stationary vehicles} is limited, the detours intersect and need to be smoothed to satisfy the kinematic constraints of the autonomous vehicle, as shown in Fig.\ref{fig_environment_segmentation}(c).

\textcolor{blue}{For clarity, we temporarily add distinct footnotes to variables in this subsection, with $left$ and $right$ representing the variables associated with the \textcolor{blue}{stationary vehicles} positioned on the left and right, respectively. 
Fig.\ref{fig_environment_segmentation}(c) illustrates these variables, with footnote $left$ in orange and $right$ in purple. }
The arc $P_1 P_2$ is part of $middle$ derived from the corner point \textcolor{blue}{$(p_{x2,left}, p_{y2,left})$} of the \textcolor{blue}{stationary vehicle} on the left, while $P_2 P_3$ is $plus$ derived from the same point. 
Meanwhile, the arc $Q_1 Q_2$ is $minus$ derived from the corner point \textcolor{blue}{$(p_{x1,right},p_{y1,right})$} of the \textcolor{blue}{stationary vehicle} on the right, while $Q_2 Q_3$ is part of $middle$ derived from the same point.

Another turn $K_1 K_2$ centered at \textcolor{blue}{$(o_{x4},o_{y4})$} with the same radius of $R$ is required so that the two \textcolor{blue}{stationary vehicles} can be connected smoothly. The center \textcolor{blue}{$(o_{x4},o_{y4})$} satisfies:
\begin{equation}
  \begin{split}
    & (o_{x3, left} - o_{x4})^2 + (o_{y3,left}-o_{y4})^2 = 4R^2 \\ 
    & (o_{x3, right} - o_{x4})^2 + (o_{y3, right}-o_{y4})^2 = 4R^2
  \end{split}
\end{equation}
Then, \textcolor{blue}{$(o_{x4},o_{y4})$} can be solved as: 
\begin{equation}
  \begin{split}
    & o_{x4} = o_{x3,left} + 2Rcos\alpha \\
    & o_{y4} = o_{y3,left} + 2Rsin\alpha
  \end{split}
\end{equation}
where, $\alpha = arcsin(-\frac{b}{4R}+\frac{\sqrt{a^2[a^2+b^2-(\frac{a^2+b^2}{4R})^2]}}{a^2+b^2})$, $a=o_{x3, left}-o_{x3, right}$, and $b=o_{y3,left}-o_{y3,right}$.

\begin{figure*}[htb]
	\centering
    \includegraphics[width=7.1in]{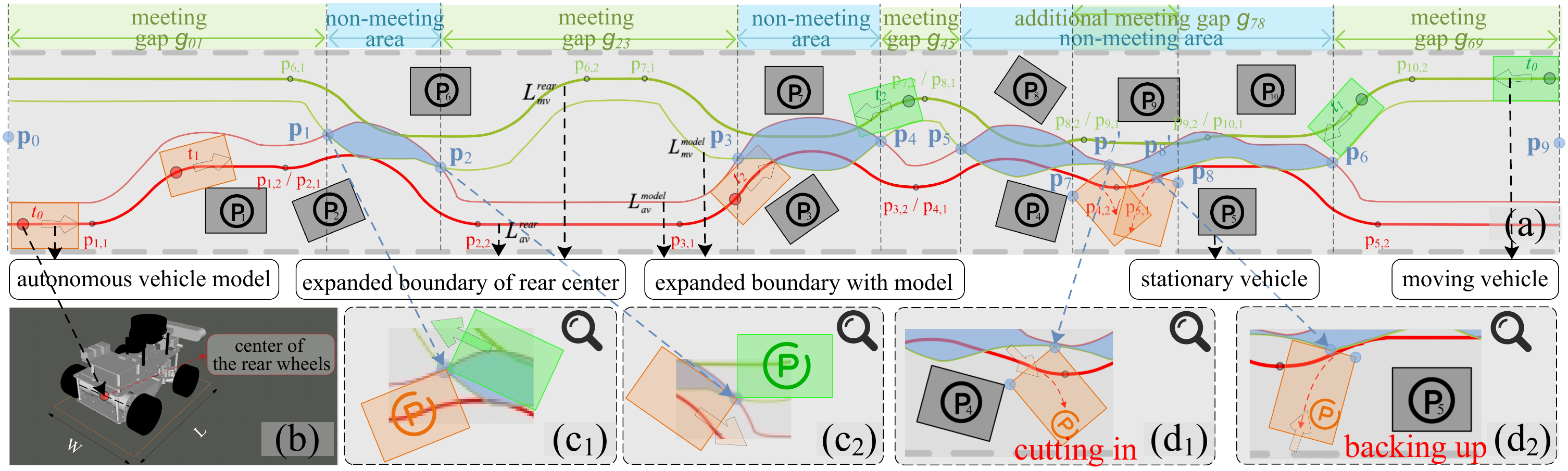}
	\caption{\textcolor{blue}{Illustration of idenfitying candidate meeting gaps. }
    (a) An expanded boundary of the rear center of the autonomous vehicle, $L_{av}^{rear}$, is obtained by connecting detours around \textcolor{blue}{stationary vehicles} $L_{av}^{rear}=\{p_{1,1} \to p_{1,2}\} \cup \{p_{2,1} \to p_{2,2}\} \cup ... \{ p_{5,1} \to p_{5,2}\}$. Expanding $L_{av}^{rear}$ according to the vehicle model, another expanded boundary with the autonomous vehicle's model $L_{av}^{model}$ is obtained. Apply the same process to the \textcolor{blue}{moving vehicle}, $L_{mv}^{rear}$ and $L_{mv}^{model}$ are obtained. Non-meeting areas are identified where a point on $L_{av}^{model}$ is higher than that on $L_{mv}^{model}$. 
    (b) The model of the autonomous vehicle. 
    (c$_1$-c$_2$) When the autonomous vehicle meets the \textcolor{blue}{moving vehicle} near the non-meeting area $\textbf{p}_1 \Rightarrow \textbf{p}_2$, the autonomous vehicle must wait before $\textbf{p}_1$ until the \textcolor{blue}{moving vehicle} advances out of the non-meeting area, or the \textcolor{blue}{moving vehicle} must wait before $\textbf{p}_2$ until the autonomous vehicle advances out of it. 
    (d$_1$-d$_2$) Two additional maneuvers of cutting in and backing up introduce an additional meeting gap $g_{78}$, enabling a more efficient meeting. 
    }
	\label{fig_extract_gaps}
\end{figure*}

\subsection{Identification of Meeting Gaps and Non-Meeting Areas} \label{3E}
It is critical to identify meeting gaps wide enough for two vehicles to pass each other on a narrow road. In the first three subsections, an \textit{expanded boundary of the rear center of the autonomous vehicle}, $L_{av}^{rear}$, can be obtained by smoothly connecting all detours around \textcolor{blue}{stationary vehicles}. Following the same principle, an \textit{expanded boundary of the rear center of the \textcolor{blue}{moving vehicle}} approaching from the opposite direction, $L_{mv}^{\text{rear}}$, can also be obtained. The two boundaries are illustrated in Fig.\ref{fig_extract_gaps}(a) as thick red and green curves, assuming the traffic rule requires vehicles to keep to the right.

Expand every point on $L_{av}^{rear}$ and $L_{mv}^{rear}$ by the vehicle model with width W and length L (see Fig.\ref{fig_extract_gaps}b), \textit{expanded boundary with the autonomous vehicle's model} $\mathcal{L}_{av}^{model}$ and \textit{expanded boundary with the \textcolor{blue}{moving vehicle's model}} $\mathcal{L}_{mv}^{model}$ can be obtained, as shown in Fig.\ref{fig_extract_gaps}(a-b) with thin red and green curves respectively. 
For each two points $P_{av}=(x_,y_{av})$ on $\mathcal{L}_{av}^{model}$ and $P_{mv}=(x,y_{mv})$ on $\mathcal{L}_{mv}^{model}$ with the same $x$ coordinate, 
if $y_{av}>y_{mv}$, the vehicles cannot pass each other at the current $x$, designating it a non-meeting area. 
Otherwise, $x$ belongs to a meeting gap, where the vehicles can pass each other smoothly. 
At the top bar of Fig.\ref{fig_extract_gaps}(a), non-meeting areas are marked in blue, and meeting gaps are marked in green.

In simple cases, the autonomous vehicle can wait at the end of the meeting gap until the \textcolor{blue}{moving vehicle} moves out of the non-meeting area (like waiting within the gap $g_{01}$ in Fig.\ref{fig_extract_gaps}c$_1$), or the \textcolor{blue}{moving vehicle} waits in place until the autonomous vehicle moves out of the non-meeting area (like waiting within the gap $g_{23}$ in Fig.\ref{fig_extract_gaps}c$_2$). 
In such cases, the two vehicles can easily meet with only two simple maneuvers: advancing and stopping.

In more complex scenarios, such as a non-meeting area  $\textbf{p}_5 \Rightarrow \textbf{p}_6$ in Fig.\ref{fig_extract_gaps}(a), there is no meeting gap for a long distance. 
But another two maneuvers of cutting in and backing up (they will be detailed later) can provide additional meeting gaps. 
As shown in Fig.\ref{fig_extract_gaps}(d$_1$), a cutting-in maneuver allows the autonomous vehicle to move closer to the road boundary at $p_{4,2}$ or even deeper, thus sparing more space for the \textcolor{blue}{moving vehicle}. 
During the cutting-in, the autonomous vehicle model interacts with $L_{mv}^{model}$ at $\textbf{p}_7'$, and $\textbf{p}_7$ is a corner point closest to the left-\textcolor{blue}{stationary vehicle}. 
Similarly, as shown in Fig.\ref{fig_extract_gaps}(d$_2$), a backing-up maneuver allows the autonomous vehicle to interact with $L_{mv}^{model}$ at $\textbf{p}_8'$, and $\textbf{p}_8$ is a corner point closest to the right-\textcolor{blue}{stationary vehicle}. 
These two maneuvers introduce an additional meeting gap $g_{78}$.

\subsection{Evaluation and Selection of Optimal Meeting Gap} \label{3F}
\begin{figure}[htb]
	\centering
    \includegraphics[width=3.4in]{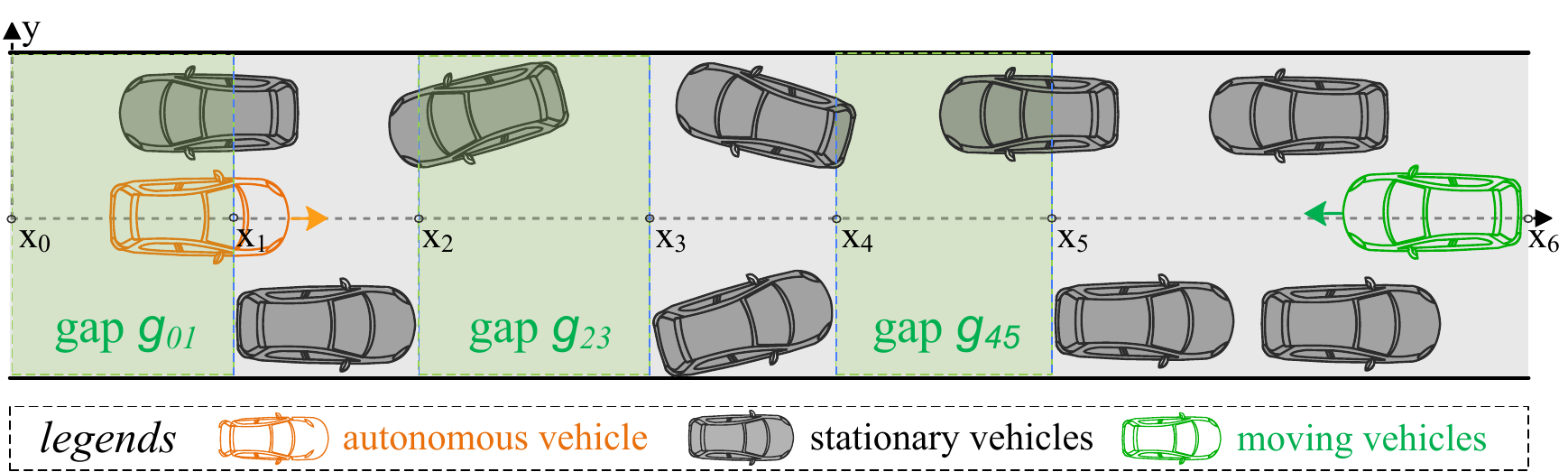}
	\caption{\textcolor{blue}{Illustration of the simplified model on the narrow road scenario.} 
    }
	\label{fig_optimal_gap}
\end{figure}

In a narrow road scenario, although there are many vehicles and meeting gaps, from the perspective of an autonomous vehicle, a simplified model consists only of itself, the closest moving vehicle, and at most three gaps, as illustrated in Fig.\ref{fig_optimal_gap}.

Based on the last subsections, three candidate meeting gaps $\mathbb{G}=\{g_{01}, g_{23}, g_{45}\}$ can be identified. 
Assuming that the autonomous vehicle and the oncoming \textcolor{blue}{moving vehicle} maintain their speeds, they will meet at a position $x$ in the future. There are four possible situations: 

(1) If $x \in [x_2,x_3]$ or $x \in [x_4,x_5]$, there is no conflict between the two vehicles, and the gap $g_{23}$ or $g_{45}$ will be selected as optimal $g^*$. 
(2) If $x \in [x_3,x_4]$, since this area cannot accommodate two vehicles, the autonomous vehicle must either cautiously slow down to $g_{23}$ or bravely accelerate to $g_{45}$. The cost for each gap is a weighted sum of four items: the gap length $c_{len}$, \textcolor{blue}{the times selecting the gap in the last ten decisions $c_{times}$}, the distance between meeting position $x$ and the gap $c_{dis}$, and whether the gap allows the vehicle to follow the lane direction $c_{lane}$. 
The gap with the lowest cost is selected as the optimal gap: 
\begin{equation}
    g^* = \mathop{\arg\min}\limits_{g \in \mathbb{G} } \textcolor{blue}{\alpha}(- w_1 c_{len} - \textcolor{blue}{w_2 c_{times}} + w_3 c_{dis} - w_4 c_{lane} ) \label{cost}
\end{equation}
\textcolor{blue}{where $\alpha$ is 0.9 if the gap was selected at the last decision and 1 otherwise, to enhance decision robustness.} 
(3) If $x \in [x_1,x_2]$, the autonomous vehicle needs to reverse into $g_{01}$ or accelerate into $g_{23}$. The items considered are the same as the Equ.(\ref{cost}), except that the cost of the reversing maneuver is multiplied by a discount factor that is less than 1 to encourage safety. 
(4) If $x > x_5$, the autonomous vehicle will move to the gap $g_{45}$.

\section{Decision-making for Autonomous Vehicle} \label{sec:4}
\textcolor{blue}{Given a meeting gap, this section initializes candidate trajectories, optimizes them under constraints, and selects the optimal one. During this process, the concept of the homology class is introduced to initialize trajectories with distinct meanings and ensure these meanings remain consistent throughout the optimization.}

\subsection{Trajectory Initialization with Homology Class Attribution} \label{4A}

There are a total of 7 candidate strategies/trajectories for each meeting gap: three that guide the autonomous vehicle to move alongside its \textcolor{blue}{lane direction}, three alongside the \textcolor{blue}{oncoming lane direction}, and one that guides the autonomous vehicle to move alongside the road centerline.

\begin{figure}[htb]
	\centering
    \includegraphics[width=3.5in]{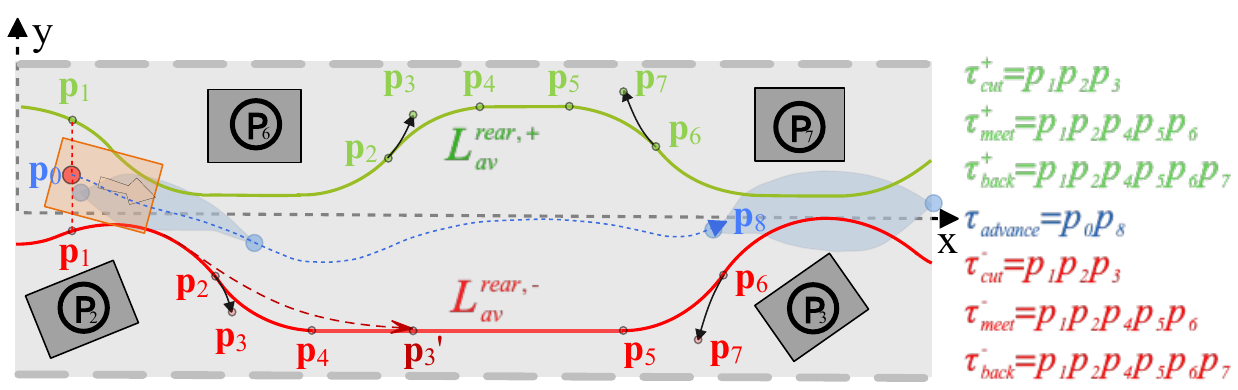}
	\caption{Illustration of 7 trajectories belonging to different homology classes. 
    }
	\label{trajectories}
\end{figure}

Focusing on the trajectories alongside the \textcolor{blue}{lane direction} of the autonomous vehicle, four key points $\textbf{p}_2,\textbf{p}_3,\textbf{p}_6,\textbf{p}_7$ are illustrated in Fig.\ref{trajectories}. 
A switch point $\textbf{p}_2$ transitions the autonomous vehicle from the state of moving alongside the \textcolor{blue}{stationary vehicle} to the state of cutting in. 
The vehicle can continue forward until it approaches the road boundary at an endpoint $\textbf{p}_3$. 
Another switch point $\textbf{p}_6$ transitions the vehicle from the state of moving alongside the \textcolor{blue}{stationary vehicle} to the state of backing up. 
The vehicle can continue reversing until it approaches the road boundary at an endpoint $\textbf{p}_7$. 

Based on these four points, a strategy \textcolor{blue}{\textit{Cut in by the lane direction} $\tau_{cut}^-$} consists of three phases. 
The first phase starts from the autonomous vehicle's current position $\textbf{p}_0$ to its nearest point $\textbf{p}_1$ on the expanded boundary of the rear center $\mathcal{L}_{av}^{rear, -}$. \textcolor{blue}{(In this section, we temporarily use the superscript ``-'' to denote the boundary alongside the autonomous vehicle’s lane direction and ``+'' to denote the boundary alongside the oncoming lane direction.)} 
The second phase continuously follows $\mathcal{L}_{av}^{rear, -}$ until the switch point $\textbf{p}_2$. 
The third phase maintains the orientation at $\textbf{p}_2$ until reaching the endpoint $\textbf{p}_3$. 
This strategy is mainly used for quick avoidance of an aggressive oncoming \textcolor{blue}{moving vehicle}.
Another trajectory \textcolor{blue}{\textit{Meet by the lane direction} $\tau_{meet}^-$} consists of two phases. 
The first phase mirrors the trajectory $\textbf{p}_0\textbf{p}_1$. 
The second phase continuously follows $\mathcal{L}_{av}^{rear, -}$ until the switch point $\textbf{p}_6$. 
This strategy is primarily used when the autonomous vehicle and the oncoming vehicle reach an agreement.
A third trajectory, \textcolor{blue}{\textit{Back up by the lane direction} $\tau_{back}^-$}, extends $\tau_{meet}^-$ by adding a third phase, where the vehicle reverses along the curve from $\textbf{p}_6$ until the endpoint $\textbf{p}_7$.

\textcolor{blue}{The primary difference between these three types of trajectories lies in the quadrant where their endpoints are located: the fourth, first, and third quadrants, respectively. 
In fact, they have a more academic definition, as they belong to different homology classes. The criterion for determining whether two trajectories belong to the same homology class is derived from ~\cite{H_signature}, which states:
\textit{
Two trajectories \( \tau_1 \) and \( \tau_2 \), connecting the same starting point and two endpoints $\textbf{e}_1$ and $\textbf{e}_2$, respectively, 
are homologous if and only if $\tau_1 \sqcup \textbf{e}_1 \textbf{e}_2 \sqcup -\tau_2$ forms the complete boundary of a $2D$ manifold embedded in the configuration space \( C \) that does not contain any stationary vehicles, and the orientations of the two endpoints are in the same quadrant. 
}
This implies that, within this gap, two trajectories $\textbf{p}_1\textbf{p}_3'$, with the endpoint $\textbf{p}_3'$ oriented in the fourth quadrant, 
and $\textbf{p}_1\textbf{p}_2\textbf{p}_3$, with the endpoint $\textbf{p}_3$ also oriented in the fourth quadrant, form a manifold $\textbf{p}_1\textbf{p}_2\textbf{p}_3\textbf{p}_3'\textbf{p}_1$ 
that does not include any stationary vehicles. When this condition is satisfied, the two trajectories belong to the same homology class and can be transformed into each other during subsequent optimization.
}

Following the same principle, the autonomous vehicle can move alongside another road boundary $\mathcal{L}_{av}^{rear,+}$ according to the oncoming lane direction, as shown in Fig.\ref{trajectories}, forming three trajectories \textcolor{blue}{\textit{Cut in by the oncoming lane direction} $\tau_{cut}^+$}, \textcolor{blue}{\textit{Meet by the oncoming lane direction} $\tau_{meet}^+$}, and \textcolor{blue}{\textit{Back up by the oncoming lane direction} $\tau_{back}^+$}. 

\textcolor{blue}{When there is no oncoming vehicle, the autonomous vehicle can \textcolor{blue}{\textit{normally advance alongside the road centerline} $\tau_{advance}$} whose points consist of the midpoint of the two expanded boundaries $\mathcal{L}_{av}^{rear,-}$ and $\mathcal{L}_{av}^{rear,+}$.}

\subsection{Constrained Optimization for Trajectories} \label{4B}
\textcolor{blue}{The seven candidate trajectories are optimized to eliminate redundant turns, avoid collisions, and satisfy the kinematic constraints of the autonomous vehicle, while ensuring that each trajectory remains within its original homology class.}

\textcolor{blue}{Assuming a candidate trajectory $\tau$ consisting of $n$ waypoints $\textbf{p}_i=(x_i,y_i,\theta_i)_{i\in\{0,1,\cdots,n-1\}}$, all trajectories share the same objective function to minimize the time to the goal~\cite{2016_TEB, graphicTEB}}: 
\begin{equation}
    \tau^* = \mathop{\arg\min}\limits_{\tau} \sum_{t=1}^{n-1} ||\textbf{p}_i-\textbf{p}_{i-1}||_2 / v_{max} \label{objective}
\end{equation}
where \( v_{max} \) is the maximum speed of the autonomous vehicle. The constraints of the optimizations for different trajectories are demonstrated separately below.

\textcolor{blue}{\textbf{Normally advance alongside the road centerline} $\tau_{advance}$: 
This strategy works when there are no oncoming moving vehicles or the autonomous vehicle resumes a normal driving state after the oncoming vehicle has passed the autonomous vehicle waiting in the gap.} 

\textcolor{blue}{The first constraint of this strategy is the location of the guidance goal $\mathcal{G}(\textbf{p}_{n-1})$ that the last waypoint $\textbf{p}_{n-1}$ of the trajectory should achieve.} The guidance goal $(x_g, y_g)$, located a few meters ahead in the lane direction of the autonomous vehicle and updated in real time, is defined as the midpoint between the two expanded boundaries $\mathcal{L}_{av}^{rear,-}$ and $\mathcal{L}_{av}^{rear,+}$, guiding the autonomous vehicle's forward movement: 
\begin{equation}
\mathcal{G}(\textbf{p}_{n-1}) = \left \{ 
\begin{array}{l}
x_{n-1} = x_g \\
y_{n-1}= \frac{\mathcal{L}_{av}^{rear,-}(x_g) + \mathcal{L}_{av}^{rear,+}(x_g)}{2} \\
\theta_{n-1} = 0 
\end{array}
\right.
\end{equation}

\textcolor{blue}{The second constraint is the autonomous vehicle's kinematics $\mathcal{H}(\textbf{p}_i,\textbf{p}_{i+1})$}, which require the curvature of adjacent waypoints to remain consistent~\cite{2017_TEB}. For any adjacent waypoints: 
\begin{equation}
    \mathcal{H}(\textbf{p}_i,\textbf{p}_{i+1}) = \begin{bmatrix} cos\theta_i + cos\theta_{i+1} \\ sin\theta_i + sin\theta_{i+1} \\ 0 \end{bmatrix} \times \begin{bmatrix} x_{i+1} - x_{i} \\ y_{i+1}-y_{i} \\ 0 \end{bmatrix} = 0
    \label{kinematics}
\end{equation}

\textcolor{blue}{The third constraint is avoiding collision with stationary vehicles $\mathcal{O}_{stationary}(\textbf{p}_i)$}. Any waypoint $\textbf{p}_i$ is required to locate between two expanded boundaries of the rear center of the autonomous vehicles $\mathcal{L}_{av}^{rear,-}$ and $\mathcal{L}_{av}^{rear,+}$: 
\begin{equation}
\mathcal{O}_{stationary}(\textbf{p}_i) = \left \{ 
\begin{array}{l}
y_i >= \mathcal{L}_{av}^{rear,-}(x_i) \\
y_i <= \mathcal{L}_{av}^{rear,+}(x_i)
\end{array}
\right.
\label{avoid_stationary}
\end{equation}

\textcolor{blue}{\textbf{Meet by the lane direction} $\tau_{meet}^-$: 
This strategy is the one most commonly adopted in narrow road meeting scenarios. 
Within the optimal meeting gap, it allows the autonomous vehicle to move forward as much as possible toward the critical safe meeting point $\textbf{p}_6 = (x_6, y_6, \theta_6)$ that is illustrated in Fig.\ref{trajectories}, thereby improving the forward efficiency. }

\textcolor{blue}{The first constraint is the location of the guidance point $\textbf{p}_6$.} To ensure that its homology attribute remains unchanged, the trajectory's orientation should stay within the first quadrant, enabling the autonomous vehicle to restart quickly: 
\begin{equation}
\mathcal{G}(\textbf{p}_{n-1}) = \left \{ 
\begin{array}{l}
x_{n-1} = x_6 \\
y_{n-1} = y_6 \\
\theta_6 \leq \theta_{n-1} \leq \frac{\pi}{2}
\end{array}
\right.
\label{g_meet}
\end{equation}

\textcolor{blue}{The second and third constraints are the autonomous vehicle's kinematics $\mathcal{H}(\textbf{p}_i, \textbf{p}_{i+1})$ and avoidance of stationary vehicles $\mathcal{O}_{stationary}(\textbf{p}_i)$}, as described in Equ.(\ref{kinematics}) and (\ref{avoid_stationary}). The only difference is that $\mathcal{O}_{stationary}(\textbf{p}_i)$ ignores the final waypoint $\textbf{p}_{n-1}$, as it has been defined in Equ.(\ref{g_meet}). 

\textcolor{blue}{The fourth constraint is the avoidance of moving vehicles $\mathcal{O}_{moving}$.} This constraint is necessary for two reasons. First, the objective function in Equ.(\ref{objective}) aims for time-optimality, driving the autonomous vehicle to move toward the guidance point along the shortest route at maximum speed. This behavior prevents the vehicle from moving next to the road edge, causing a collision risk with oncoming \textcolor{blue}{moving vehicles}. Second, most of the time, the meeting occurs during the autonomous vehicle's movement toward \( \textbf{p}_6 \) rather than after it reaches \( \textbf{p}_6 \). 
The constraint $\mathcal{O}_{moving}$ is considered in the spatiotemporal dimension. 
For a \textcolor{blue}{moving vehicle} with four corner points, each pair of adjacent points $\textbf{p}_j$ and $\textbf{p}_k$ can form a line segment \( Ax + By + C = 0 \). 
For the corner point \( \textbf{p}_i = (x_i, y_i) \) of the autonomous vehicle that is closest to the \textcolor{blue}{moving vehicle} in the spatiotemporal dimension and falls within the \textcolor{blue}{moving vehicle}'s model, \( ||Ax_i + By_i + C|| \) can be used to calculate the shortest distance between the autonomous vehicle and the \textcolor{blue}{moving vehicle}~\cite{2017_TEB}. 
The constraint $\mathcal{O}_{moving}(\textbf{p}_i)$ requires this distance to be greater than 0 to ensure collision avoidance: 
\begin{equation}
  \mathcal{O}_{moving}(\textbf{p}_i) = ||Ax_i + By_i + C|| > 0
  \label{cons_moving}
\end{equation}

\begin{figure}[htb]
	\centering
    \includegraphics[width=3.5in]{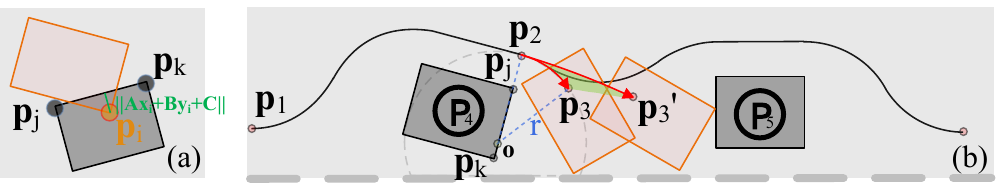}
	\caption{Illustration of (a) the constraint of avoiding moving vehicles and (b) the movable range (green) of the target during cutting in $\tau_{cut}^-$. 
    }
	\label{cutin}
\end{figure}

\textcolor{blue}{\textbf{Cut in by the lane direction} $\tau_{cut}^-$:
This strategy works when the oncoming vehicle is moving at high speed and the autonomous vehicle needs to perform a rapid evasive action to ensure safety. }

\textcolor{blue}{The guidance point constraint $\mathcal{G}(\textbf{p}_{n-1})$ of this trajectory is more flexible.} 
When the autonomous vehicle cuts into the gap, its target \( \textbf{p}_{n-1} \) adapts dynamically to the oncoming \textcolor{blue}{moving vehicles}. 
Cutting deeply allows the autonomous vehicle to move closer to the road boundary, sparing more space for \textcolor{blue}{moving vehicles}. 
As the autonomous vehicle turns around a corner point \( \textbf{p}_j = (p_{xj}, p_{yj}) \) in Fig.\ref{cutin}(b), its turning radius varies from the minimum turning radius \( R \) to \( \infty \). 
The orientation is constrained to remain within the fourth quadrant to preserve the trajectory's homology attribute. In this case:
\begin{equation}
\mathcal{G}(\textbf{p}_{n-1}) = \left \{ 
\begin{array}{l}
(x_{n-1}-o_{x})^2 + (y_{n-1}-o_{y})^2 \geq R^2  \\
\frac{3\pi}{2} \leq \theta_{n-1} \leq 2\pi
\end{array}
\right.
\label{g_cut}
\end{equation}
where $(o_{x}, p_{y})$ is the circle center around corner point $\textbf{p}_j$, $o_{x} = p_{xj} - (R-\frac{W}{2}) \frac{p_{xj} - p_{xk}}{||\textbf{p}_j-\textbf{p}_k||_2}$ and $o_{y} = p_{yj} - (R-\frac{W}{2}) \frac{p_{yj} - p_{yk}}{||\textbf{p}_j-\textbf{p}_k||_2}$

\textcolor{blue}{The other constraints of this strategy, including $\mathcal{H}(\textbf{p}_i, \textbf{p}_{i+1})$, $\mathcal{O}_{stationary}(\textbf{p}_i)$, and $\mathcal{O}_{moving}(\textbf{p}_i)$, are the same as Equ.(\ref{kinematics}), (\ref{avoid_stationary}), and (\ref{cons_moving})}. The final waypoint $\textbf{p}_{n-1}$ is also ignored in these items, as it has been specially constrained in Equ.(\ref{g_cut}).

\textcolor{blue}{\textbf{Back up by the lane direction} $\tau_{back}^-$:
This strategy works when there is no meeting gap between the autonomous vehicle and the moving vehicle, forcing the autonomous vehicle to reverse to avoid a collision and ensuring the robustness of the meeting process. 
Another situation is that when sufficient time is available to meet at a tiny gap where $\tau_{meet}^-$ cannot be executed, $\tau_{back}^-$ enables the autonomous vehicle to advance further and restart earlier, enhancing efficiency.}

\textcolor{blue}{The guidance point constraint of this trajectory \( \mathcal{G}(\textbf{p}_{n-1}) \) is also flexible.} Backing up deeply creates more space, encouraging the oncoming moving vehicle to advance with confidence. While backing up around a corner point \( \textbf{p}_j = (p_{xj}, p_{yj}) \), the turning radius varies from \( R \) to \( \infty \), and the velocity orientation of \( \textbf{p}_{n-1} \) is constrained to remain within the third quadrant:
\begin{equation}
\mathcal{G}(\textbf{p}_{n-1}) = \left \{ 
\begin{array}{l}
(x_{n-1}-o_{x})^2 + (y_{n-1}-o_{y})^2 \geq R^2  \\
\pi \leq \theta_{n-1} \leq \frac{3\pi}{2}
\end{array}
\right.
\label{g_back}
\end{equation}
where $(o_{x}, p_{y})$ is the circle center around the corner point $\textbf{p}_j$.

\textcolor{blue}{The other constraints of this strategy, including $\mathcal{H}(\textbf{p}_i, \textbf{p}_{i+1})$, $\mathcal{O}_{stationary}(\textbf{p}_i)$, and $\mathcal{O}_{moving}(\textbf{p}_i)$, are the same as Equ.(\ref{kinematics}), (\ref{avoid_stationary}), and (\ref{cons_moving})}. The final waypoint $\textbf{p}_{n-1}$ is also ignored in these items, as it has been specially constrained in Equ.(\ref{g_back}). 

\textcolor{blue}{When turning by the oncoming lane direction, the strategies \textbf{meet by the oncoming lane direction} $\tau_{meet}^+$, \textbf{cut in by the oncoming lane direction} $\tau_{cut}^+$, and \textbf{back up by the oncoming lane direction} $\tau_{back}^+$ follow the same principle. They are not elaborated further here.}

\subsection{Evaluation and Selection for Trajectories} \label{4C}
\textcolor{blue}{The optimal trajectory $\tau^*$ is selected from the candidates $\{ \tau_{advance}, \tau_{meet}^+, \tau_{cut}^+, \tau_{back}^+, \tau_{meet}^-, \tau_{cut}^-, \tau_{back}^- \}$ through a hierarchical evaluation process.}

\textcolor{blue}{The first layer checks for an oncoming moving vehicle.} If none is detected or the meeting is completed, the strategy $\tau_{advance}$ is selected to normally move forward. 

\textcolor{blue}{The second layer handles lane selection when an oncoming moving vehicle is present}. If the oncoming vehicle actively moves to a lane, the autonomous vehicle takes the opposite lane using $\{ \tau_{meet}^+, \tau_{cut}^+, \tau_{back}^+ \}$ or $\{ \tau_{meet}^-, \tau_{cut}^-, \tau_{back}^- \}$. If the oncoming vehicle doesn't show its preferred lane, the gap with more available space determines the autonomous vehicle's choice. When the available space on the autonomous vehicle’s lane is larger, the optimal strategy will be selected among $\{\tau_{meet}^+, \tau_{cut}^+, \tau_{back}^+\}$; otherwise, the optimal strategy will be selected among $\{\tau_{meet}^-, \tau_{cut}^-, \tau_{back}^-\}$.

\textcolor{blue}{The third layer identifies the optimal trajectory.} The smooth meeting strategy $\tau_{meet}^+$ or $\tau_{meet}^-$ is always preferred. If unavailable, such as the gap $g_{78}$ in Fig.\ref{fig_extract_gaps}, the backing-up strategy $\tau_{back}^+$ or $\tau_{back}^-$ will be preferred for efficient movement. If the relative distance between the two vehicles is insufficient for the autonomous vehicle to complete the time-consuming backing-up maneuver, the cut-in strategy $\tau_{cut}^+$ or $\tau_{cut}^-$ is adopted to ensure safety.

\section{Simulation Results} \label{sec:5}
\textcolor{blue}{This section first introduces the simulation platform, including the robot model, scenarios, evaluation metrics, and comparison methods.
Next, qualitative and quantitative analyses are conducted on four challenging scenarios to validate the safety, efficiency, and robustness of the proposed SM-NR.}

\subsection{Simulation Setup}
There are three types of vehicles in the simulation: the autonomous vehicle, \textcolor{blue}{stationary vehicles}, and oncoming \textcolor{blue}{moving vehicles}, marked in orange, black, and green rectangular boxes, respectively, in the following figures. 
The size of the autonomous vehicle is [0.26 m $\times$ 0.186 m], with a maximum speed of 0.5 m/s. 
The road length is 7.0 m, and the width is 0.92 m. An SL coordinate system is established, as shown in Fig.\ref{normal_simulation}(a). 
The autonomous vehicle moves along the positive x-axis direction from \( x = 0 \) to \( x = 7 \), while the oncoming moving vehicles move in the opposite direction, from \( x = 7 \) to \( x = 0 \).
\textcolor{blue}{During this process, the autonomous vehicle's localization algorithm uses the open-source AMCL algorithm\footnote{https://github.com/ros-planning/navigation}. 
The detection of road boundaries, \textcolor{blue}{stationary vehicles}, and \textcolor{blue}{moving vehicles} positions is provided by the simulation platform with Gaussian noise added to the ground truth data. 
The estimation of the current and future speeds of moving vehicles is performed using the Kalman filter algorithm\footnote{https://github.com/koide3/hdl\_people\_tracking}.} 
Visit the source code at \href{https://sm-nr.github.io}{https://sm-nr.github.io} for more details. 
\textcolor{blue}{As for the moving vehicles, in the qualitative experiments for each scenario, they are manually controlled by volunteers to interact with the autonomous vehicle's actions, allowing for an in-depth test of the algorithm's robustness.}

\textcolor{blue}{In the quantitative comparison, the trajectories of the moving vehicles are replays of historical data. This historical data records the trajectories of moving vehicles when different volunteers controlled both the autonomous and moving vehicles to perform smooth meetings. This setup ensures that the moving vehicles do not react to the autonomous vehicle during testing, making all collisions attributable solely to the autonomous vehicle's lack of avoidance ability, thereby guaranteeing fairness in the comparison.
The stationary vehicles in the data recording are randomly distributed, with orientations varying within \([-20^\circ, 20^\circ] \cup [160^\circ, 200^\circ]\), and their distances from the road centerline exceeding 0.15 m to ensure that the stationary vehicles are positioned near the road boundary.}

In the four test scenarios, each method is tested 25 times with different historical data. 
\textcolor{blue}{During this process, five metrics are evaluated: success rate $s$, travel time $t$, time ratio $r_{time}$, decision oscillation ratio $r_{dec}$, and decision frequency $f_{dec}$.
The success rate $s$ is defined as the number of times the autonomous vehicle reaches \( x = 7 \) without a collision, divided by the total 25 test runs, focusing on safety. 
Travel time $t$ refers to the average time taken by the autonomous vehicle to move from the starting point to the endpoint, focusing on efficiency. 
The time ratio $r_{time}$ is the ratio of the additional time caused by avoiding oncoming moving vehicles, calculated as: 
\begin{equation}
  r_{time} = \frac{t_{\text{with moving vehicles}} - t_{\text{without moving vehicles}}}{t_{\text{without moving vehicles}}}
\end{equation}
The decision oscillation ratio $r_{dec}$ is the ratio of decision changes (e.g., switching optimal gaps or transitioning from cut-in to back-up strategies) within 10 consecutive decisions to the total 10 decisions, focusing on decision robustness. 
Decision frequency $f_{dec}$ refers to the number of decisions made by the autonomous vehicle per second, focusing on real-time responsiveness.}
Examples of the calculation for these metrics can be found in the following subsections.

\textcolor{blue}{Two methods are compared in the simulation test: Timed Elastic Band (TEB)~\cite{2017_TEB,graphicTEB} and P2EG~\cite{IROS_ZHANG_2022}. TEB is a widely used motion planning method in autonomous navigation. It samples candidate trajectories using a probabilistic map and also introduces the concept of homology classes to differentiate trajectories. However, TEB only considers trajectories based on clockwise or counterclockwise detours around other vehicles, without utilizing the unique characteristics of the narrow meeting problem, such as gap identification and the detailed distinction of strategies like cut-ins and backups.
P2EG is a method tailored for narrow road meetings. It identifies candidate meeting gaps by simply evaluating spatial margin. Using a lattice planner~\cite{lattice}, it samples multiple candidate target points meters ahead of the vehicle and fits candidate trajectories with polynomials. It then employs a tentative game to model vehicle interactions and reach a mutually accepted solution. However, its gap evaluation considers only spatial size and overlooks the kinematics of vehicles, leading to some candidate meeting gaps being ignored.}

\subsection{Performance in Single Normal Meeting Gap Scenario}
\begin{figure}[thb]
    \centering
    \includegraphics[width=3.4in]{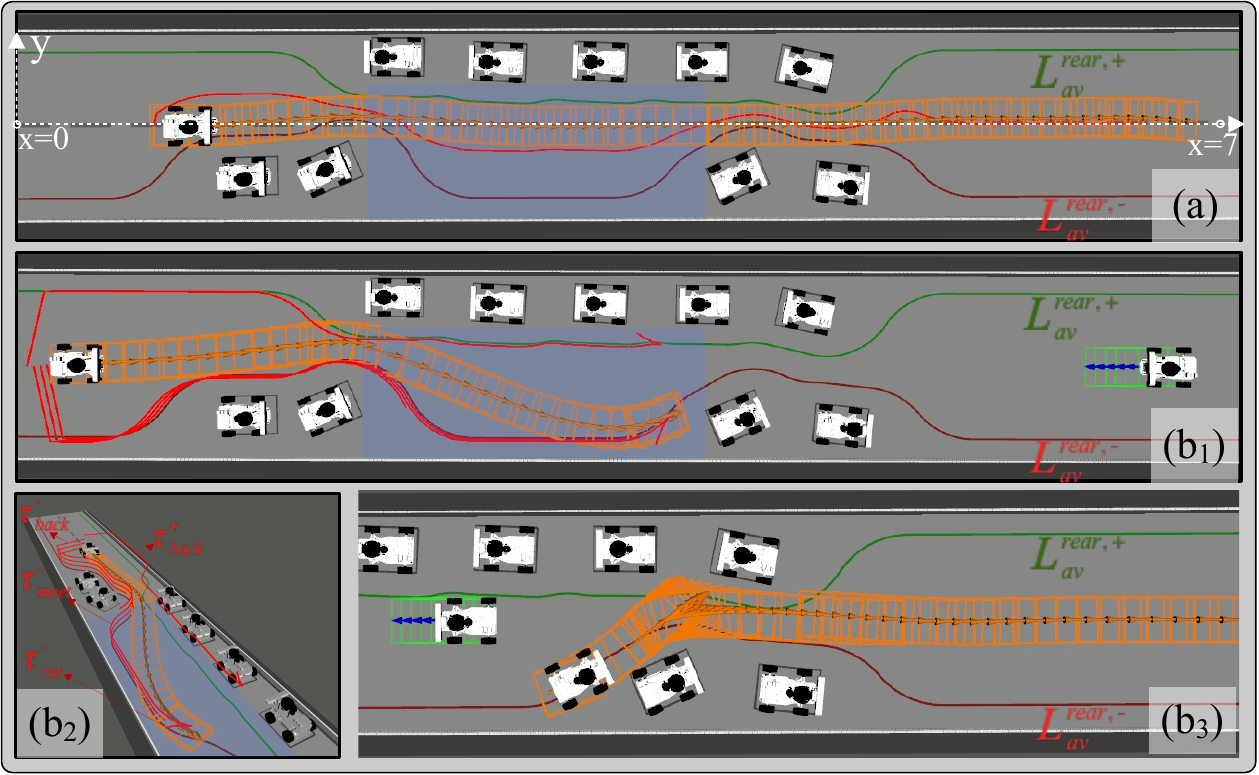}
    \caption{Illustration of the single normal meeting gap scenario. (a) When there is no moving vehicle, the autonomous normally advances by the strategy $\tau_{advance}$ in orange. 
    (b$_1$) When there is an oncoming moving vehicle, the autonomous vehicle has four strategies shown in stark red and picks $\tau^-_{meet}$. 
    (b$_2$) illustrates four strategies with a clear perspective. 
    (b$_3$) After the meeting, the autonomous vehicle resumes the normal advance with strategy $\tau_{advance}$. 
    }
    \label{normal_simulation}
\end{figure}

\textcolor{blue}{This scenario, abbreviated as \textit{\textbf{single}} and illustrated in Fig.\ref{normal_simulation}, represents the simplest meeting scenario with only one large gap. 
It provides a detailed explanation of the autonomous vehicle's motion process and the metric time ratio \( r_{time} \).}

\textcolor{blue}{In Fig.\ref{normal_simulation}(a), when there is no oncoming vehicle, the autonomous vehicle normally advances alongside the road centerline with the strategy \( \tau_{advance} \) which is marked in red.} Its points are the midpoints of \( L_{av}^{rear,-} \) and \( L_{av}^{rear,+} \). 
The optimization process stretches this trajectory \( \tau_{advance} \) into a near-straight line, eliminating redundant curves to improve efficiency. 
\textcolor{blue}{The optimized trajectory is visualized with orange arrows and executed by the autonomous vehicle to move from \( x=0 \) to \( x=7 \) with a time consumption \( t_{\text{without moving vehicle}} \)}.

\textcolor{blue}{In Fig.\ref{normal_simulation}(b$_1$-b$_3$), as a comparison, when there is an oncoming moving vehicle, the autonomous vehicle needs to park at a gap for avoidance.} 
Considering the relative distance and speed of the two vehicles, the gap marked in blue in Fig.\ref{normal_simulation}(b$_1$) is selected. 
Four strategies, including \textit{meet by the lane direction} \( \tau_{meet}^- \), \textit{cut in by the lane direction} \( \tau_{cut}^- \), \textit{back up by the lane direction} \( \tau_{back}^- \), and \textit{back up by the oncoming lane direction} \( \tau_{back}^+ \), are visualized in Fig.\ref{normal_simulation}(b$_2$) from a clear perspective. 
Note that strategies \( \tau_{meet}^+ \) and \( \tau_{cut}^+ \) are ignored in simulation and experiment as they are rarely used in meeting problems.

In this case, \( \tau_{meet}^- \) is selected as the optimal trajectory, guiding the autonomous vehicle to park at the gap until the moving vehicle passes. 
\textcolor{blue}{Afterward, as shown Fig.\ref{normal_simulation}(b$_3$), the autonomous vehicle resumes normal driving navigation with the strategy \( \tau_{advance} \), targeting a point meters ahead.} 
\textcolor{blue}{The entire time consumption for moving from \( x=0 \) to \( x=7 \) is recorded as \( t_{\text{with moving vehicle}} \). Now the time ratio \( r_{time} \) can be calculated directly, reflecting the additional time caused by avoiding the oncoming vehicle}.

\begin{figure*}[htb]
	\centering
    \includegraphics[width=7.1in]{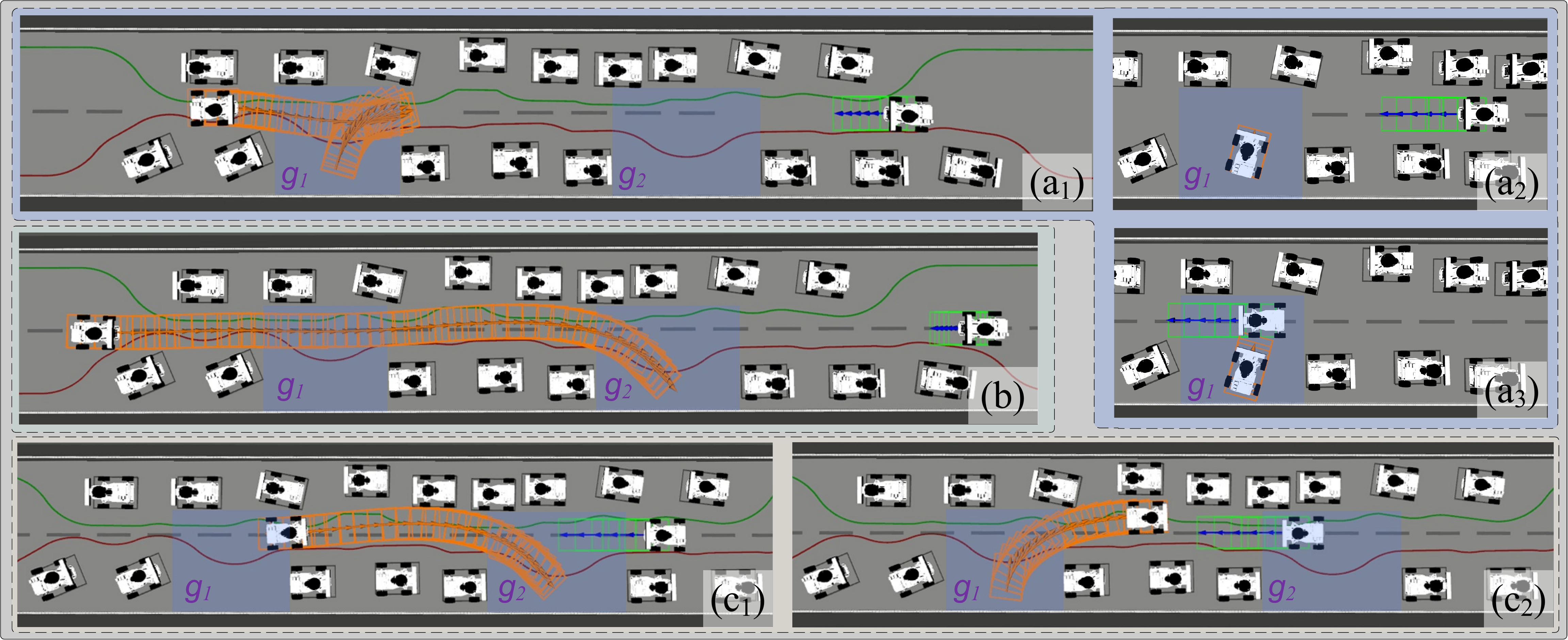}
	\caption{Illustration of the robustness and efficiency in a conflict scenario with two meeting gaps. 
    (a$_1$) For an aggressive oncoming moving vehicle, the autonomous vehicle backs up at the closer gap $g_1$. 
    (a$_2$-a$_3$) Upon detecting that the moving vehicle does not cooperatively move alongside stationary vehicles, the autonomous vehicle backs up further for safety. 
    (b) For a moderate moving vehicle, the autonomous vehicle bravely cuts into the farther gap $g_2$ for efficiency. 
    (c$_1$-c$_2$) For a moving vehicle that suddenly accelerates, the autonomous vehicle robustly switches from the farther gap $g_2$ to the closer gap $g_1$.
    }
	\label{game_simulation}
\end{figure*}
\subsection{Performance in Conflict Scenario with Two Meeting Gaps}

\textcolor{blue}{This complex scenario, abbreviated as \textit{\textbf{conflict}} and illustrated in Fig.\ref{game_simulation}, requires the autonomous vehicle to select the optimal gap from two options. It evaluates the robustness and efficiency using four metrics: decision frequency \( f_{dec} \), decision oscillation ratio \( r_{dec} \), travel time \( t \), and time ratio \( r_{time} \).}

\textcolor{blue}{In Fig.\ref{game_simulation}(a$_1$-a$_3$), when the oncoming vehicle moves from \( x=7 \) to \( x=0 \) at a higher speed, the autonomous vehicle compromises and selects the closer gap \( g_1 \). 
Since this gap is too narrow to execute \( \tau_{meet}^- \), the strategy \( \tau_{back}^- \) is chosen.} 
To optimize efficiency for the subsequent restart, the trajectory is initialized to back up slightly, leaving minimal passing space for the oncoming vehicle, as shown in Fig.\ref{game_simulation}(a$_1$).
However, this minimal space increases the risk of collision, particularly when the oncoming vehicle does not proactively shift right to avoid the autonomous vehicle, as illustrated in Fig.\ref{game_simulation}(a$_2$). 
In this situation, Equ.(\ref{g_back}) enables the target point \( \textbf{p}_{n-1} \) of \( \tau_{back}^- \) to adjust adaptively, allowing the autonomous vehicle to back up further and ensure a safer meeting, as shown in Fig.\ref{game_simulation}(a$_3$).

\textcolor{blue}{In Fig.\ref{game_simulation}(b), for an oncoming vehicle with a lower speed, the autonomous vehicle chooses the gap \( g_2 \) to minimize unnecessary waiting time and restart earlier. Considering the relative positions and speeds, there is insufficient time for \( \tau_{meet}^- \) or \( \tau_{back}^- \), so \( \tau_{cut}^- \) is selected to avoid collision quickly.}

\textcolor{blue}{In Fig.\ref{game_simulation}(c$_1$-c$_2$), a unique situation arises when the oncoming vehicle initially compromises to meet at gap \( g_1 \) but then suddenly accelerates. 
Our approach, SM-NR, enables the autonomous vehicle to quickly switch its decision from gap \( g_1 \) to gap \( g_2 \) for safety. 
The autonomous vehicle backs up to gap \( g_2 \) and waits until the moving vehicle passes by.}

\textcolor{blue}{This \textit{\textbf{conflict}} scenario tests the robustness of our approach against prediction errors and decision frequency.} 
The autonomous vehicle's decisions remain consistent regarding the optimal gap and strategy (\( \tau_{cut}^- \) or \( \tau_{back}^- \)), even in the presence of perception errors for the \textcolor{blue}{moving vehicle}'s position or speed, like in scenarios of Fig.\ref{game_simulation}(a-b). 
\textcolor{blue}{This robustness is attributed to the design of Equ.(\ref{cost}).} 
The items \( c_{len} \) and \( c_{dis} \) dominate the initial decision, leading to the selection of the most appropriate gap. 
In the following decisions, even with prediction noise and a slight change of \( c_{dis} \), \( c_{times} \) and \( \alpha \) ensure that the previously chosen gap continues to have a lower cost, thus maintaining decision robustness. 
\textcolor{blue}{Two metrics, decision oscillation ratio \( r_{dec} \) and decision frequency \( f_{dec} \), shown in Table.\uppercase\expandafter{\romannumeral1}, validate this robustness.}
As observed, \( r_{dec} \) is nearly 0 across all scenarios, with minor oscillations arising from adjustments to the oncoming vehicle's behavior or rare instances of significant prediction errors, which are promptly corrected in subsequent decisions. 
In contrast, P2EG demonstrates noticeable decision oscillations, mainly due to frequent changes in the optimal gap.
\begin{figure}[thb]
	\centering
    \includegraphics[width=3.4in]{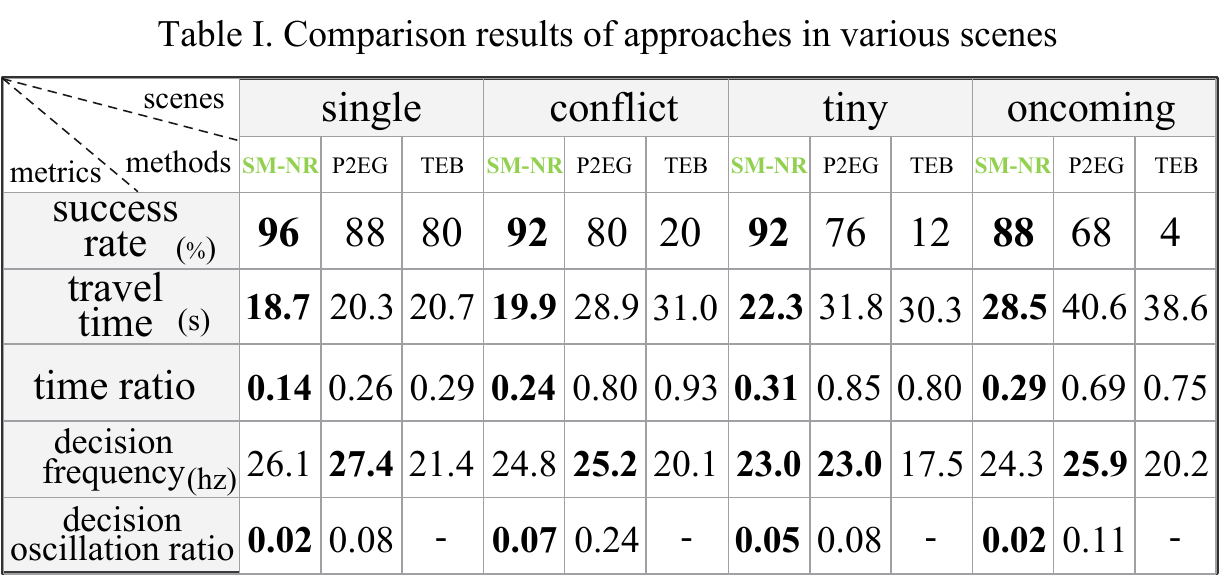}
	\label{table}
\end{figure}
\textcolor{blue}{Additionally, our SM-NR achieves a competitively high decision frequency \( f_{dec} \) of approximately 24 Hz on a laptop with Intel i7-12600h, enabling the autonomous vehicle to respond quickly to changes in oncoming moving vehicles. This capability, as demonstrated in the scenario depicted in Fig.\ref{game_simulation}(c), enhances robustness in complex situations.} 

\textcolor{blue}{Regarding efficiency, the flexibility of the guidance point constraints in \( \tau_{cut}^- \) (Equ.\ref{g_cut}) and \( \tau_{back}^- \) (Equ.\ref{g_back}) allows the endpoint of the trajectory to adapt dynamically within the gap. This adaptability minimizes unnecessary entry into the gap, enabling the autonomous vehicle to resume normal navigation more quickly. 
In contrast, TEB lacks such adaptability within gaps, leading to less efficient navigation. As shown in Table.\uppercase\expandafter{\romannumeral1}, SM-NR achieves significantly lower time ratios \( r_{time} \) and travel times \( t \), demonstrating its superior efficiency.}

\begin{figure*}[htb]
	\centering
    \includegraphics[width=7.1in]{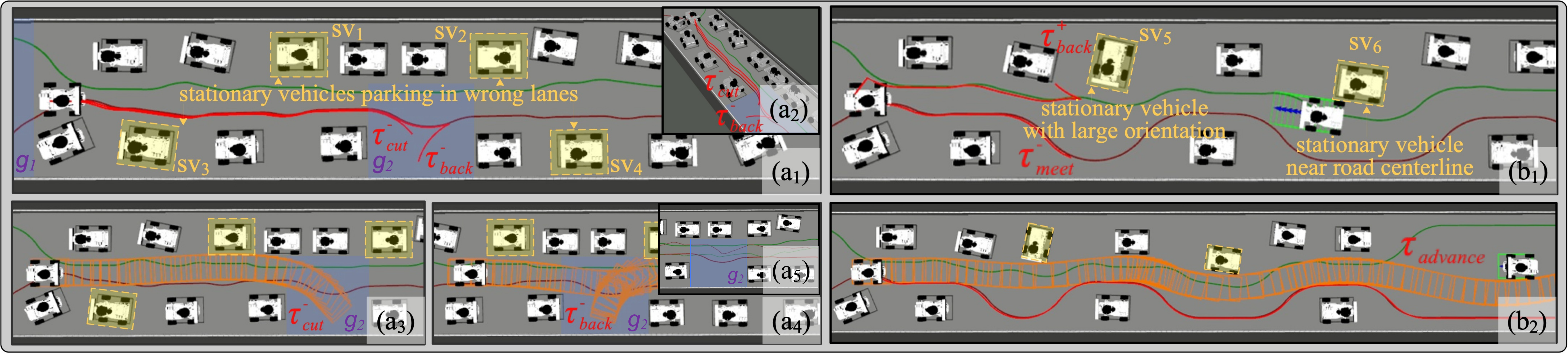}
	\caption{    
    \textcolor{blue}{Illustration of the tiny scenario.} By modeling stationary vehicles as corner points, the expanded boundaries and strategies can be robustly identified, even in challenging situations such as (a$_1$-a$_5$) stationary vehicles parked in the wrong lane or (b$_1$) stationary vehicles with large orientations or positioned near the road centerline. 
    (a$_3$, a$_4$, b$_2$) When gaps and roads are extremely narrow, the trajectory can still be safely generated and executed.    
    }
	\label{tiny_simulation}
\end{figure*}
\subsection{Performance in Tiny Scenario with Irregular Vehicles}
\textcolor{blue}{This very narrow scenario, abbreviated as \textit{\textbf{tiny}} and illustrated in Fig.\ref{tiny_simulation}, validates the safety, efficiency, and robustness under challenging conditions with narrower roads and gaps, including stationary vehicles parked in the wrong lane, with large orientations, or positioned near the road centerline. The evaluation metrics, success rate \( s \), travel time \( t \), and time ratio $r_{time}$ highlight the advantages of SM-NR.}

\textcolor{blue}{In Fig.\ref{tiny_simulation}(a$_1$), several stationary vehicles \{\( sv_1 \), \( sv_2 \), \( sv_3 \), \( sv_4 \)\} are irregularly parked in the wrong lane with opposite orientations. By modeling each stationary vehicle as four corner points, SM-NR ensures vehicle orientation does not affect expanded boundary generation or trajectory initialization. 
As shown in Fig.\ref{tiny_simulation}(a$5$), the gap $g_2$ is too narrow to be identified by P2EG or SM-NR only using $\tau_{meet}^-$. However, our designed $\tau_{cut}^-$ and $\tau_{back}^-$, shown in Fig.\ref{tiny_simulation}(a$_3$) and (a$_4$), enable the robust identification of $g_2$, allowing the autonomous vehicle to advance more to park at $g_2$ before meeting, thereby improving efficiency. 
As shown in Table.\uppercase\expandafter{\romannumeral1}, SM-NR achieves the lowest travel time of 22.3~s and time ratio of 0.31. outperforming P2EG and TEB, which inefficiently force the vehicle to \( g_1 \).}

\textcolor{blue}{In Fig.\ref{tiny_simulation}(b$_1$), another two irregularly spaced stationary vehicles \{\( sv_5 \), \( sv_6 \)\} are presented, with \( sv_5 \) parked at a large orientation and \( sv_6 \) near the road centerline. 
By modeling stationary vehicles as corner points, the expanded boundary around \( sv_5 \) and the strategy \( \tau_{back}^+ \) are generated robustly. When the oncoming vehicle moves slowly, the strategy \( \tau_{advance} \) allows the autonomous vehicle to navigate smoothly forward, as shown in Fig.\ref{tiny_simulation}(b$_2$). 
Regarding safety, constraints based on the expanded boundary ensure high safety even in narrow gaps at Fig.\ref{tiny_simulation}(a$_3$-a$_4$) or narrow roads leaving only minimal safety margins outside the autonomous vehicle’s width at Fig.\ref{tiny_simulation}(b$_2$). As shown in Table \uppercase\expandafter{\romannumeral1}, SM-NR achieves a success rate of 92\%, significantly surpassing other methods in terms of safety.}

\subsection{Performance in Meeting by Oncoming Lane Directions}
\begin{figure}[thb]
	\centering
    \includegraphics[width=3.4in]{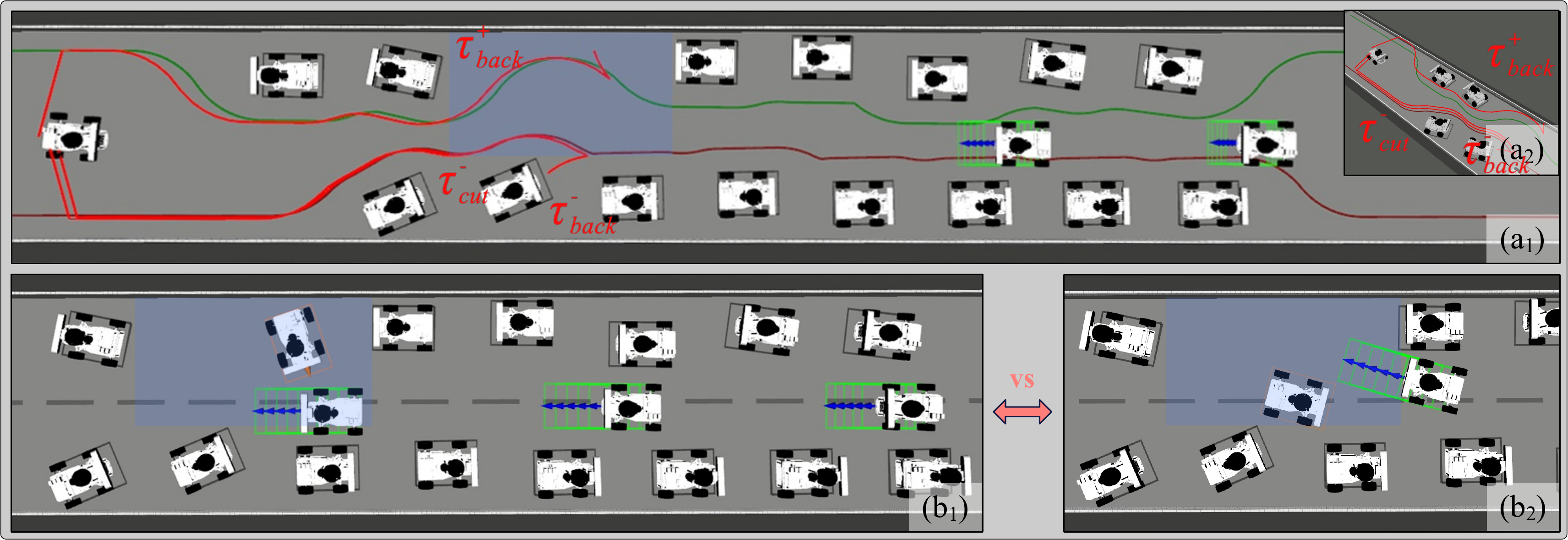}
	\caption{Illustration of the meeting by oncoming lane direction. (a$_1$) When the gap is located on the oncoming lane, (a$_2$) the autonomous vehicle has three options: \( \{\tau^+_{back}, \tau^-_{back}, \tau^-_{cut}\} \). 
    (b$_1$) Backing up on the oncoming lane \( \tau^+_{back} \) frees up more space for oncoming vehicles, while (b$_2$) parking on the autonomous vehicle's lane places higher demands on the oncoming drivers.
    }
	\label{left_simulation}
\end{figure}

\textcolor{blue}{This scenario, abbreviated as \textit{\textbf{oncoming}} and illustrated in Fig.\ref{left_simulation}, is a unique case where the meeting gap is located on the oncoming lane of the autonomous vehicle. It demonstrates the robustness and safety of our approach by introducing the strategy \( \tau^+_{back} \) through ablation experiments.} 

Since the gap is narrow, the strategy of normally meeting \( \tau^-_{meet} \) is not feasible. Therefore, the autonomous vehicle has three candidate strategies: \textit{back up by the lane direction} \( \tau^-_{back} \), \textit{cut in by the lane direction} \( \tau^-_{cut} \), and \textit{back up by the oncoming lane direction} \( \tau^+_{back} \), as shown in Fig.\ref{left_simulation}(a$_1$-a$_2$).

\textcolor{blue}{In Fig.\ref{left_simulation}(b$_1$), when there are multiple oncoming vehicles moving at low speeds, the autonomous vehicle selects the strategy \( \tau^+_{back} \) for an earlier restart.} Realizing that the oncoming vehicles do not cooperatively shift, the autonomous vehicle backs up deeper, thereby creating more space for the oncoming moving vehicles to pass safely.
\textcolor{blue}{This symmetric design on the oncoming lane is a key factor contributing to our method's robustness and safety in complex scenarios. 
This is reflected quantitatively, as shown in Table.\uppercase\expandafter{\romannumeral1}, where our SM-NR achieves a significantly higher success rate of 88\%, compared to P2EG's 68\% and TEB's 4\%.} 

\textcolor{blue}{Additionally, qualitative ablation analysis further validates the importance of this design.} When the autonomous vehicle employs the strategy \( \tau^+_{back} \) to back up into the gap, oncoming vehicles, controlled by volunteers, almost always navigate around without collisions. In contrast, \textcolor{blue}{removing \( \tau^+_{back} \) forces the autonomous vehicle to park in its lane direction using \( \tau^-_{cut} \) or \( \tau^-_{back} \). As illustrated in Fig.\ref{left_simulation}(b$_2$), executing \( \tau^-_{cut} \) leaves minimal margin for oncoming vehicles, significantly increasing the likelihood of collisions with the autonomous vehicle or stationary vehicles, even when volunteers operate the oncoming vehicles with caution.}

\section{Real-world Experiment} \label{sec:6}


\subsection{Experiment Setup} \label{sec:71}
\textcolor{blue}{This section verifies the feasibility of applying the proposed method to physical vehicles.} 
The vehicle models used in the real-world experiments are identical to those in the simulated environment. Both the autonomous vehicle and the \textcolor{blue}{oncoming moving vehicle} are Ackermann models with dimensions [0.26 m $\times$ 0.186 m] and a maximum speed of 0.5 m/s. 
\textcolor{blue}{(Due to source limitations of our lab, a real autonomous vehicle is unavailable.)} 
The road is 4.0 m long and 1.0 m wide, with \textcolor{blue}{stationary vehicles} represented by cardboard boxes measuring [0.30 m $\times$ 0.20 m]. 
The autonomous vehicle is equipped with an Orin Nano motherboard running Ubuntu 18.04, and localization is performed using the AMCL algorithm. The average decision-making frequency is approximately 17 Hz.

\textcolor{blue}{Since the proposed approach focuses on decision-making and motion planning, and there is no open-source vehicle detection and tracking algorithm with high accuracy available, two techniques are employed to mitigate interference from detection and tracking errors (which is another widely recognized challenge in the industry). 
First, the positions of the \textcolor{blue}{stationary vehicles} are preset and provided to the algorithm with real-time Gaussian noise added (mean = 0, variance = 0.05) to simulate detection errors. 
Second, the odometer and motherboard of the oncoming moving vehicle can calculate its position and velocity in real-time. This information is also augmented with Gaussian noise and shared with the autonomous vehicle via WIFI+ROS, simulating vehicle detection results.}

\textcolor{blue}{In the experiment, the vehicles approaching from the oncoming side of the autonomous vehicle are controlled by volunteers. These volunteers operate the vehicle from a bird's-eye view, standing outside the scene and overseeing the entire environment. 
To ensure familiarity and minimize operational errors, each volunteer is given over 5 minutes of free control to practice maneuvering the vehicle before the test, and each volunteer is allowed to perform 5 trials for each scenario to gain experience and ensure consistency in their actions.}

\begin{figure*}[htb]
	\centering
    \includegraphics[width=7.0in]{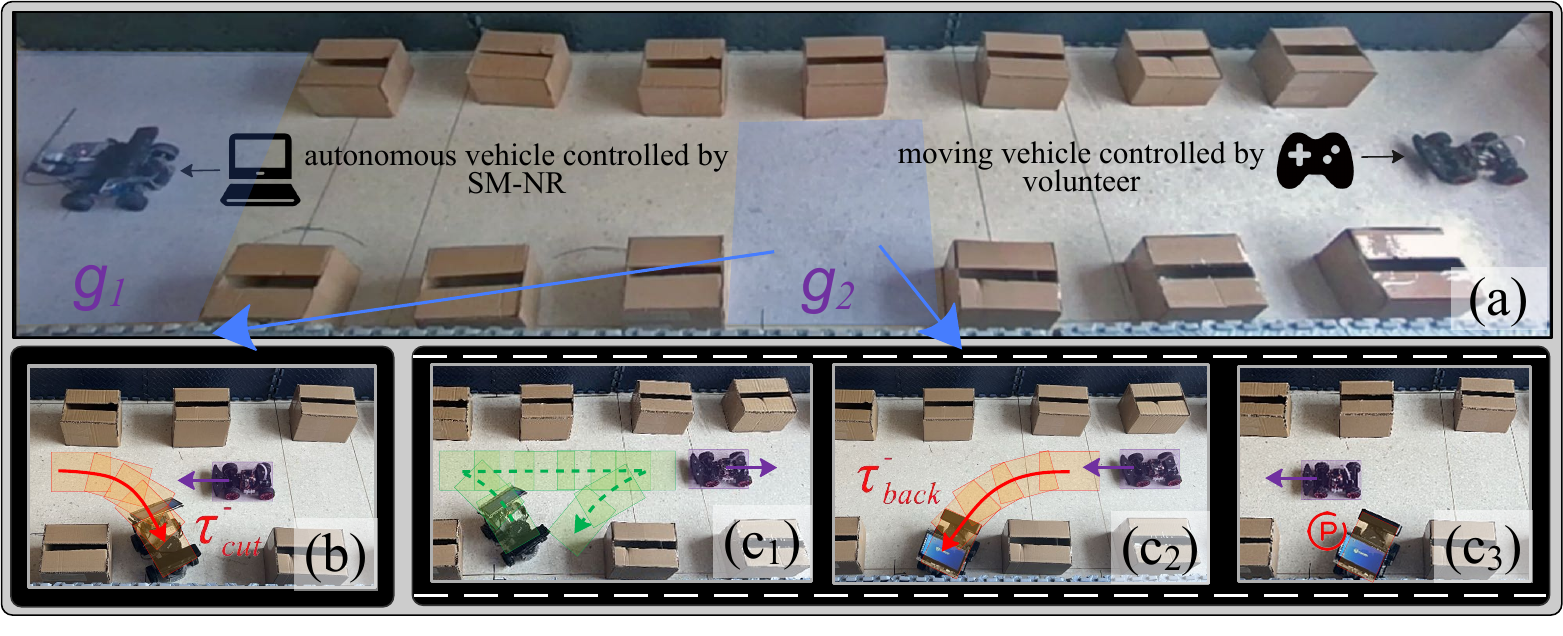}
	\caption{Performance of the proposed SM-NR in the tiny meeting gap scenario. 
    (a) The SM-NR-controlled vehicle moves from the left end and passes the oncoming vehicle to the right end. 
    (b) SM-NR guides the vehicle to cut in the gap when the oncoming moving vehicle moves fast. 
    (c$_1$-c$_3$) For another situation, when the oncoming vehicle moves backwards due to the fear of collision, our SM-NR enables the autonomous vehicle to change its strategy to back up deeper into the gap, leaving more movable space to encourage the oncoming vehicle to move forward.
    }
	\label{exp_tiny}
\end{figure*}
\subsection{Performance in Tiny Scenario} \label{sec:72}
\textcolor{blue}{Extending from the \textit{tiny} scenario in simulation, this subsection validates the robustness of our approach in the presence of volunteers exhibiting aggressive or cautious behavior, as shown in Fig.\ref{exp_tiny}.}
In this scenario, the autonomous vehicle starting from the left end is controlled by SM-NR, while the vehicle starting from the right end is controlled by a volunteer. The narrowness of the scenario lies in the small distances between adjacent \textcolor{blue}{stationary vehicles} on the same lane, allowing only two strategies: \textit{cut in by the lane direction} $\tau^-_{cut}$ and \textit{back up by the lane direction} $\tau^-_{back}$. 

\textcolor{blue}{As shown in Fig.\ref{exp_tiny}(b), observing the fast movement of the oncoming vehicle controlled by an aggressive volunteer, SM-NR directs the autonomous vehicle to adopt the strategy $\tau^-_{cut}$ to avoid collision.} To facilitate a quicker restart, it leaves only the space of the oncoming moving vehicle's size instead of cutting directly into the road edge. The aggressive volunteer advances the oncoming vehicle, prompting the autonomous vehicle to cut in deeper for collision avoidance.

\textcolor{blue}{As a comparison, Fig.\ref{exp_tiny}(c$_1$-c$_3$) illustrates the meeting process with a more cautious volunteer. Initially, SM-NR also directs the autonomous vehicle to adopt the strategy $\tau^-_{cut}$ to avoid collisions. Due to concerns about potential collisions, the volunteer hesitates to advance. Recognizing sufficient time and space, SM-NR switches the autonomous vehicle from $\tau^-_{cut}$ to $\tau^-_{back}$ to enable a faster restart, as shown in Fig.\ref{exp_tiny}(c$_1$).} This backing-up maneuver signals the volunteer of the autonomous vehicle's yielding intention, encouraging the volunteer to move forward confidently while the autonomous vehicle backs up further, achieving a safe meeting, as shown in Fig.\ref{exp_tiny}(c$_2$-c$_3$).

\textcolor{blue}{Additionally, 10 more experiments have been conducted where volunteers replace SM-NR to control the vehicle starting from the left end.} An interesting phenomenon emerges: due to the narrowness, 4 volunteers hesitate to meet at the tiny gap $g_2$ and park on the leftmost side $g_1$ until the oncoming vehicle passes. Another 4 volunteers attempt to cut into gap $g_2$ but collide with \textcolor{blue}{stationary vehicles}, the wall, or the oncoming vehicle. \textcolor{blue}{Only 2 volunteers successfully meet the oncoming vehicle at the efficient meeting gap $g_2$. In contrast, SM-NR achieves collision-free meetings in 9 out of 10 tests. 
This highlights the significance of autonomous navigation and the advantages of our SM-NR in narrow-road meeting scenarios. Limited by driving ability, accurate perception of safe distances, and other drawbacks, human drivers often hesitate to advance for fear of collisions, leading to inefficiencies or traffic blockages. In contrast, autonomous navigation precisely perceives the environment and models the relationships between vehicles, enabling safe, smooth, and efficient meetings.}

\begin{figure*}[thb]
	\centering
    \includegraphics[width=7.0in]{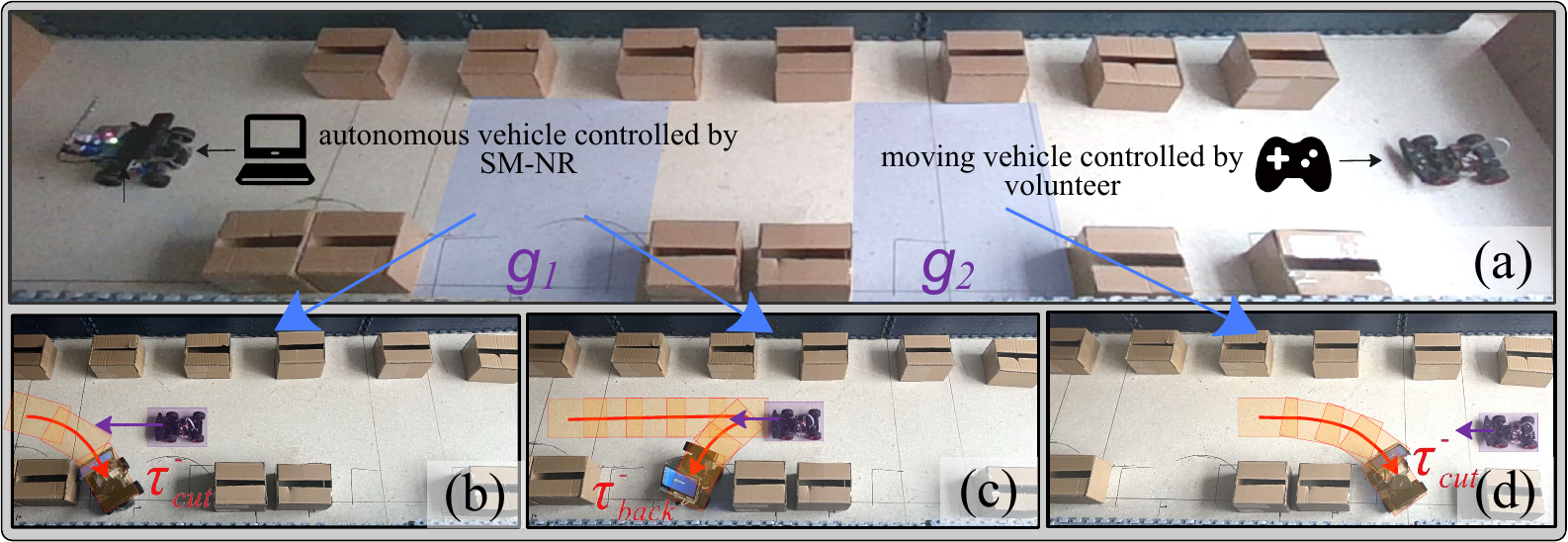}
	\caption{Performance of SM-NR in the conflict scenario with two meeting gaps. 
    (a) There are two gaps between vehicles. 
    In the order of decreasing speed of the oncoming vehicle, the autonomous vehicle 
    (b) cuts in the nearer gap, 
    (c) backs up into the nearer gap, 
    (d) or cuts in the gap away from it.
    }
	\label{exp_game}
\end{figure*}
\subsection{Performance in Conflict Scenario}
\textcolor{blue}{Extending from the \textit{conflict} scenario in simulation, this subsection validates SM-NR on a real vehicle and demonstrates its robust decision-making to avoid oncoming moving vehicles with varying behaviors, as shown in Fig.\ref{exp_game}.} 
In this scenario, two gaps are present: $g_1$, closer to the autonomous vehicle, and $g_2$, farther away. 
Depending on the behaviors of the oncoming vehicle, the autonomous vehicle evaluates its relative relationship and selects different gaps and strategies.

For an aggressive oncoming vehicle with high speed, as illustrated in Fig.\ref{exp_game}(b), the autonomous vehicle compromises by quickly cutting into the nearer gap $g_1$ using the strategy $\tau_{cut}^-$ to ensure safety. 
For a moderately moving oncoming vehicle with relatively lower speed, as shown in Fig.\ref{exp_game}(c), the autonomous vehicle also selects the nearer gap $g_1$ but prefers to back up into the gap using the strategy $\tau_{back}^-$, as there is sufficient time to execute this more time-consuming maneuver. 
For a gentle oncoming vehicle with the lowest speed, as illustrated in Fig.\ref{exp_game}(d), the autonomous vehicle bravely advances to the farther gap $g_2$ and ensures safety by cutting into it using the strategy $\tau_{cut}^-$.

\section{Conclusion and \textcolor{blue}{Discussion}} \label{sec:8}
This paper presents a refined model for navigating narrow road scenarios, enabling autonomous vehicles to traverse such conditions with both safety and efficiency. Our approach, SM-NR, introduces the principle of road width occupancy minimization to identify candidate meeting gaps. The optimal meeting gap is selected based on the relative positioning of the autonomous and oncoming vehicles. Candidate trajectories are then initialized, optimized, and evaluated using homology classes and road information. Simulations and experiments validate SM-NR in challenging scenarios, demonstrating its ability to identify more meeting gaps, efficiently advance before encounters with oncoming vehicles, safely generate candidate trajectories, and robustly select the optimal strategy.

\textcolor{blue}{However, this work has some limitations that could be addressed in future studies. First, SM-NR assumes straight road boundaries, but narrow road meetings in curved or turning scenarios may also occur and need consideration. Second, some detection tricks are employed to mitigate perception errors, but robust motion planning under perception errors, often referred to as risk-aware motion planning, remains an open and valuable research direction. Third, due to laboratory resource limitations, experiments are conducted on small-scale vehicles with Ackermann steering models, and testing SM-NR on real autonomous vehicles could uncover additional insights.}


\bibliographystyle{IEEEtran}
\bibliography{./bibtex/New_IEEEtran_how-to}

\end{document}